\documentclass[sigconf,screen]{acmart}
\AtBeginDocument{%
  }

\acmSubmissionID{9555} 
\renewcommand\footnotetextcopyrightpermission[1]{}
\graphicspath{{figures/}}

\usepackage{CJKutf8} 
\usepackage[T1]{fontenc}
\usepackage[utf8]{inputenc}
\usepackage{subcaption}
\usepackage{bm}
\usepackage[ruled,vlined]{algorithm2e}
\usepackage{booktabs}
\usepackage{xcolor}
\definecolor{darkred}{RGB}{139,0,0}
\definecolor{royalblue}{RGB}{65,105,225}
\usepackage{multirow}
\usepackage{enumitem}
\usepackage{cleveref}
\newcommand{\zh}[1]{{\begin{CJK*}{UTF8}{gbsn}#1\end{CJK*}}}

\begin{document}

\title{RoleMAG: Learning Neighbor Roles in Multimodal Graphs}

\author{Yilong Zuo}
\affiliation{%
  \institution{Beijing Institute of Technology}
  \city{Beijing}
  \country{China}
}
\email{1120231863@bit.edu.cn}

\author{Xunkai Li}
\affiliation{%
  \institution{Beijing Institute of Technology}
  \city{Beijing}
  \country{China}
}
\email{cs.xunkai.li@gmail.com}

\author{Zhihan Zhang}
\affiliation{%
  \institution{Beijing Institute of Technology}
  \city{Beijing}
  \country{China}
}
\email{3220241443@bit.edu.cn}

\author{Ronghua Li}
\affiliation{%
  \institution{Beijing Institute of Technology}
  \city{Beijing}
  \country{China}
}
\email{lironghuabit@126.com}

\author{Guoren Wang}
\affiliation{%
  \institution{Beijing Institute of Technology}
  \city{Beijing}
  \country{China}
}
\email{wanggrbit@126.com}

\renewcommand{\shortauthors}{Zuo et al.}

\acmConference[MM '26]{The 34th ACM International Conference on Multimedia}{November 10--14, 2026}{Rio de Janeiro, Brazil}
\acmBooktitle{Proceedings of the 34th ACM International Conference on Multimedia (MM '26), November 10--14, 2026, Rio de Janeiro, Brazil}
\acmYear{2026}
\copyrightyear{2026}

\begin{abstract}
Multimodal attributed graphs (MAGs) combine multimodal node attributes with structured relations. However, existing methods usually perform shared message passing on a single graph and implicitly assume that the same neighbors are equally useful for all modalities. In practice, neighbors that benefit one modality may interfere with another, blurring modality-specific signals under shared propagation. To address this issue, we propose \textbf{RoleMAG}, a multimodal graph framework that learns how different neighbors should participate in propagation. Concretely, RoleMAG distinguishes whether a neighbor should provide shared, complementary, or heterophilous signals, and routes them through separate propagation channels. This enables cross-modal completion from complementary neighbors while keeping heterophilous ones out of shared smoothing. Extensive experiments on three graph-centric MAG benchmarks show that RoleMAG achieves the best results on RedditS and Bili\_Dance, while remaining competitive on Toys. Ablation, robustness, and efficiency analyses further support the effectiveness of the proposed role-aware propagation design. Our code is available at \href{https://anonymous.4open.science/r/RoleMAG-7EE0/}{anonymous.4open.science/r/RoleMAG}.
\end{abstract}

\ccsdesc[500]{Computing methodologies~Neural networks}
\ccsdesc[300]{Computing methodologies~Artificial intelligence}

\keywords{Multimodal Graph Learning, Multimodal Attributed Graphs, Cross-Modal Inconsistency, Role-aware Propagation, Heterophily}

\maketitle

\section{Introduction}

\begin{figure}[t]
    \centering
    \includegraphics[width=0.48\textwidth]{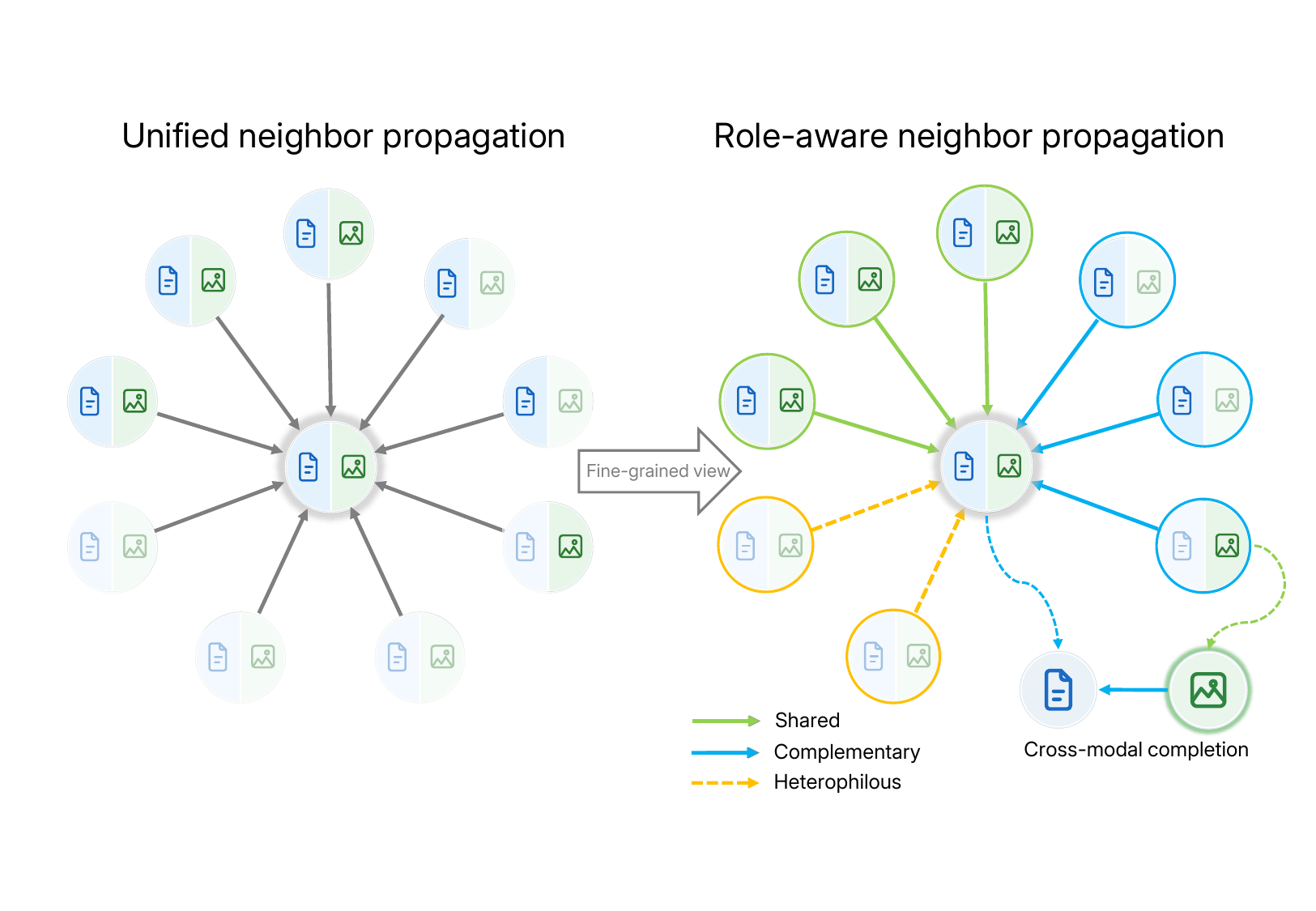}
    \caption{Illustration of the key motivation behind RoleMAG. Existing multimodal graph methods usually adopt a unified neighbor propagation pattern, implicitly treating all neighbors as if they contributed in a similar way. In practice, however, neighbors may play different roles in multimodal propagation: some provide \emph{shared} support, some serve as \emph{complementary} neighbors that enable cross-modal completion, while others correspond to \emph{heterophilous} relations that should not be mixed into shared smoothing. This motivates a fine-grained, role-aware view of neighborhood propagation.}
    \label{fig:intro}
\end{figure}

Multimodal attributed graphs (MAGs) unify graph topology with multimodal attributes such as text and images within a shared modeling framework, allowing both relational dependencies among entities and their rich semantic content to be captured simultaneously. As a result, they have become an important substrate for multimodal graph learning (MGL) in recent years~\cite{ektefaie2023multimodal,zhu2024mmgraph,wan2026openmag}. Such data arise naturally in a wide range of real-world scenarios, including recommender systems, social networks, and biochemistry~\cite{wan2026openmag,zhu2026tmte}. On the one hand, rich modality information provides semantic cues beyond pure structural modeling for graph-centric tasks such as node classification and link prediction. On the other hand, graph structure offers contextual constraints for cross-modal retrieval, alignment, and even generation, leading to representations that are more coherent and discriminative~\cite{ektefaie2023multimodal,wan2026openmag}. Consequently, how to establish more effective interactions between multimodal semantics and graph structure has become one of the central questions in MAG research.

Despite notable progress in representation learning and structural modeling, existing MAG methods still tend to adopt an overly uniform treatment during propagation. In this work, three limitations are of particular interest. \textbf{(1) \textbf{Neighbor roles are not uniform}.} Many methods perform shared message passing on a single graph, or only adjust neighborhood weights through coarse-grained dynamic mechanisms~\cite{zhu2026tmte,hong2026dip,sun2026mario}. Such designs implicitly assume that the same neighbor helps different modalities in a similar manner. In real MAGs, however, this assumption often fails. Prior studies have shown that cross-modal consistency can itself be weak, and that different nodes and their neighborhoods may exhibit clear modality preferences~\cite{sun2026mario}. \textbf{(2) \textbf{Complementary information is not explicitly modeled}.} Once neighbors are no longer treated as being ``equally useful,'' another issue emerges: even beneficial neighbors do not help in the same way. Some provide shared support, whereas others are better suited to complete missing, degraded, or weak information in a particular modality. Recent work has begun to strengthen cross-modal interaction through dynamic paths or query-based modules~\cite{hong2026dip,ning2025graph4mm,li2023blip2}. Yet in the MAG setting, which neighbors should be regarded as complementary sources, and along which direction the completion signal should flow, remain largely unmodeled. \textbf{(3) \textbf{Heterophilous relations are mixed into shared propagation}.} There is also a class of neighbors that should not be absorbed into shared smoothing. What they carry are often cross-class, cross-semantic, or high-frequency discrepant relations, rather than consistency cues that should be directly smoothed. Existing studies on heterophily have shown that low-frequency propagation alone is insufficient for such relations, and that filters capable of accommodating both homophily and heterophily are often more effective~\cite{chien2021gprgnn,bo2021fagcn}. When shared support, complementary completion, and heterophilous relations are all merged into a single propagation logic, multimodal representations are inevitably coupled too early during propagation, which compromises both discriminability and robustness.

\emph{The key question is therefore: how can different neighbor roles in multimodal propagation be identified, so that truly complementary information can complete weak modalities along appropriate cross-modal directions, while heterophilous relations are prevented from being mistakenly mixed into shared smoothing?} Building upon this observation, this work presents \underline{\textbf{RoleMAG}}. The core idea is that edges differ not only in whether they are useful, but also in how they should participate in propagation. For the same edge, the model should first determine whether it is better treated as shared support, a complementary source, or a heterophilous interaction, and then decide how it should be propagated. To this end, RoleMAG learns an edge-level distribution over three roles and routes neighborhood information into separate propagation channels accordingly. Concretely, the framework first infers the role of each edge from semantic cues, and then uses structural support to calibrate propagation strength. The three roles are defined as \emph{shared}, \emph{complementary}, and \emph{heterophilous}. Shared neighbors are then used for stable consistency propagation, complementary neighbors are passed through a directional cross-modal completion process, and heterophilous neighbors are handled by a dedicated signed filter so that they are not directly mixed into shared smoothing~\cite{hong2026dip,ning2025graph4mm,li2023blip2,chien2021gprgnn,bo2021fagcn}. Finally, the outputs of the three channels are fused into the final representation for downstream tasks.

\textbf{Our Contributions:}
\textbf{(1) \textbf{Empirical probing.}} A targeted empirical analysis is conducted to examine the diversity of neighborhood roles in real MAGs, providing evidence for the three phenomena studied in this work: neighbor utility is modality-dependent, complementary relations are often directional, and heterophilous neighborhoods should not be directly mixed into shared smoothing.
\textbf{(2) \textbf{Role-aware propagation.}} RoleMAG is introduced as an edge-level role-aware framework for multimodal graph learning. It explicitly learns how each edge is distributed over \emph{shared}, \emph{complementary}, and \emph{heterophilous} roles, and routes neighborhood information through distinct propagation channels rather than handling all neighbors with a single mechanism.
\textbf{(3) \textbf{Comprehensive evaluation.}} RoleMAG is systematically evaluated under a unified MAG benchmark, covering overall performance, mechanism effectiveness, robustness, and efficiency. The results show that the proposed method exhibits clearer advantages when neighborhood roles are more complex, while remaining competitive in the remaining settings.

\section{Empirical Study}
\label{sec:empirical_study}

Before presenting the method, three more direct questions are examined first: \textbf{(Q1)} Do neighbors contribute in the same way across modalities? \textbf{(Q2)} Is complementary propagation directional? \textbf{(Q3)} Should heterophilous neighborhoods be directly mixed into shared smoothing? To answer them, empirical analyses are conducted on three multimodal graphs, namely Grocery, Toys, and RedditS. The encoder and prediction head are kept fixed as much as possible, and only the neighborhood organization or propagation strategy is varied, so that the observed differences mainly reflect the propagation mechanism itself~\cite{wan2026openmag}.

For Q1, node classification is performed separately on the text branch and the image branch, and three neighborhood organizations are compared: the original graph $\mathcal{G}^{O}$, the text-consistent neighborhood $\mathcal{G}^{T}$, and the image-consistent neighborhood $\mathcal{G}^{I}$. Here, $\mathcal{G}^{T}$ and $\mathcal{G}^{I}$ are both formed by selecting locally consistent neighbors within the original neighborhood according to modality similarity, allowing the modality preference over neighborhood structure to be examined. For Q2, asymmetric modality degradation settings are constructed: under a fixed corruption level imposed on the target modality, the other modality is kept intact, and three settings are compared, namely symmetric aggregation, completion along the correct direction, and reverse-direction completion. For Q3, samples are grouped into low/mid/high strata according to the local heterophily ratio of each node, and two propagation strategies are compared: sending all neighbors into shared propagation, or handling heterophilous neighbors with a separate heterophily-aware propagator.

\begin{figure*}[t]
    \centering
    \begin{subfigure}[t]{0.242\textwidth}
        \centering
        \includegraphics[width=\linewidth]{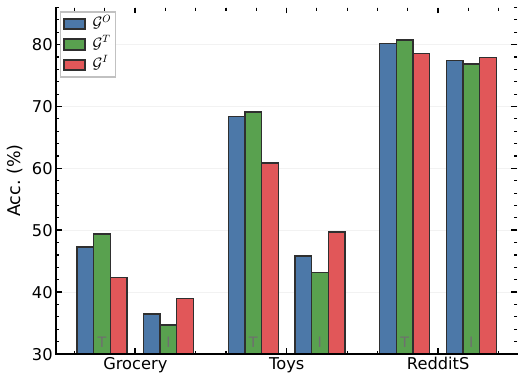}
        \caption{Modality-dependent neighbor utility.}
        \label{fig:emp_q1}
    \end{subfigure}
    \hfill
    \begin{subfigure}[t]{0.242\textwidth}
        \centering
        \includegraphics[width=\linewidth]{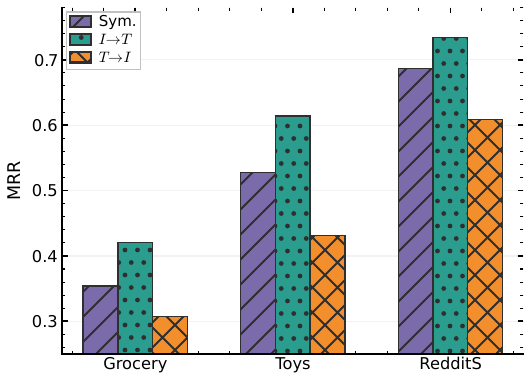}
        \caption{Directional completion when text is degraded.}
        \label{fig:emp_q2_text}
    \end{subfigure}
    \hfill
    \begin{subfigure}[t]{0.242\textwidth}
        \centering
        \includegraphics[width=\linewidth]{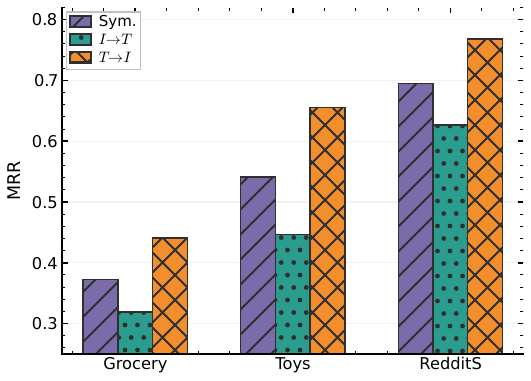}
        \caption{Directional completion when image is degraded.}
        \label{fig:emp_q2_image}
    \end{subfigure}
    \hfill
    \begin{subfigure}[t]{0.242\textwidth}
        \centering
        \includegraphics[width=\linewidth]{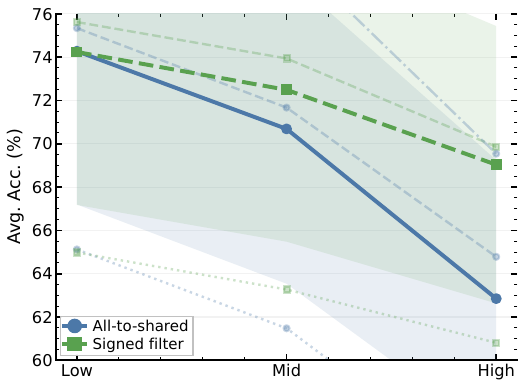}
        \caption{Heterophily-stratified propagation behavior.}
        \label{fig:emp_q3}
    \end{subfigure}

    \caption{Empirical observations behind RoleMAG. (a) Neighbor utility is modality-dependent: text and image branches prefer different neighborhood organizations. (b) When text representations are degraded, complementary completion is most effective along the $I\!\rightarrow\!T$ direction. (c) When image representations are degraded, the preferred direction reverses to $T\!\rightarrow\!I$. (d) As heterophily increases, directly mixing all neighbors into shared smoothing becomes less reliable, whereas a dedicated heterophily-aware propagator remains more stable.}
    \label{fig:empirical_study}
\end{figure*}

\textbf{Neighbors are not equivalent across modalities (Q1).}
Fig.~\ref{fig:empirical_study}(a) shows a stable pattern: the same neighborhood cannot serve both modalities equally well. For the text branch, the best results on all three datasets are obtained under the text-consistent neighborhood $\mathcal{G}^{T}$: the accuracy on Grocery improves from 47.32 on the original graph to 49.41, that on Toys rises from 68.42 to 69.11, and that on RedditS also increases slightly from 80.21 to 80.74. By contrast, once the text branch is placed on the image-consistent neighborhood $\mathcal{G}^{I}$, the performance drops noticeably on all datasets, reaching only 42.38 and 60.87 on Grocery and Toys, respectively. The image branch exhibits the opposite trend. Its best results on all three datasets are achieved under $\mathcal{G}^{I}$, with accuracies of 38.96, 49.73, and 77.91, whereas using $\mathcal{G}^{T}$ consistently causes degradation. What matters here is not merely the gain itself, but the fact that the preferred neighborhood changes with the modality. This suggests that real MAGs do not admit a single neighborhood organization that is naturally equally useful to all modalities. Under shared propagation, the model is therefore forced to average over conflicting neighborhood preferences, which weakens modality-specific signals.

\textbf{Complementary propagation is directional (Q2).}
Fig.~\ref{fig:empirical_study}(b)--(c) reveals a second phenomenon: complementary information does not act symmetrically, but depends on which modality is currently weaker. Consider first the text-degradation setting. In Fig.~\ref{fig:empirical_study}(b), completion along the $I\!\rightarrow\!T$ direction yields the best MRR on all three datasets: 0.421 on Grocery, 0.614 on Toys, and 0.734 on RedditS, all clearly higher than the 0.354, 0.528, and 0.687 obtained by symmetric aggregation. If the direction is reversed to $T\!\rightarrow\!I$, the performance further drops to 0.307, 0.431, and 0.609, respectively. When the image modality is degraded, the preferred direction reverses completely. As shown in Fig.~\ref{fig:empirical_study}(c), $T\!\rightarrow\!I$ reaches 0.441, 0.655, and 0.768 on Grocery, Toys, and RedditS, respectively, consistently outperforming symmetric aggregation and substantially surpassing the reverse direction $I\!\rightarrow\!T$. These results indicate that complementary relations are not merely about increasing the intensity of neighborhood fusion. A more specific question must be answered: \emph{which modality is currently weaker, and along which direction should completion proceed?} This also suggests that the complementary channel is better modeled as a directional path rather than an undifferentiated shared fusion branch~\cite{ning2025graph4mm,li2023blip2,jin2024instructg2i}.

\textbf{Heterophilous neighborhoods should not be directly mixed into shared smoothing (Q3).}
Fig.~\ref{fig:empirical_study}(d) complements the above findings from another angle. In low-heterophily regions, the two propagation strategies behave almost identically: averaged over datasets, the accuracies of all-to-shared and signed filtering are 74.27 and 74.24, respectively. This suggests that when the local structure is relatively consistent, sending all neighbors into shared propagation does not immediately cause obvious problems. The real divergence emerges as heterophily increases. In the mid-heterophily region, the average accuracies become 70.69 and 72.49, indicating that signed filtering has already started to show stronger stability. In the high-heterophily region, the gap further widens to 62.85 versus 69.03, a difference of 6.18 points. The same trend is clear on each dataset. In the high-heterophily group, signed filtering improves over all-to-shared by 6.58, 5.08, and 6.89 points on Grocery, Toys, and RedditS, respectively. In other words, heterophilous edges are not simply noise to be discarded. They often carry another kind of relational signal, but they are not suitable for direct absorption into shared smoothing. Once these relations are mixed with shared support, the model is more likely to wash out discriminative difference structures too early during propagation. Handling them with a dedicated heterophily-aware propagator is therefore more consistent with prior observations in heterophily graph learning~\cite{chien2021gprgnn,bo2021fagcn}.

Taken together, these findings point to the same conclusion: propagation in real MAGs is not merely about deciding which neighbors are more important, but first about determining \emph{how} they should participate in propagation. Some neighbors are better suited to provide shared support, some are more useful for directional completion, and some should be handled separately as heterophilous relations. RoleMAG is designed exactly around this observation: it identifies the role of each neighbor first, and then decides which propagation channel that neighbor should enter. This view is also aligned with recent discussions on modality preference, directional cross-modal bridging, and heterophily-aware propagation~\cite{sun2026mario,hong2026dip,ning2025graph4mm,li2023blip2,jin2024instructg2i,chien2021gprgnn,bo2021fagcn}.

\section{Methodology}

\begin{figure*}[t]
    \centering
    \includegraphics[width=0.998\textwidth]{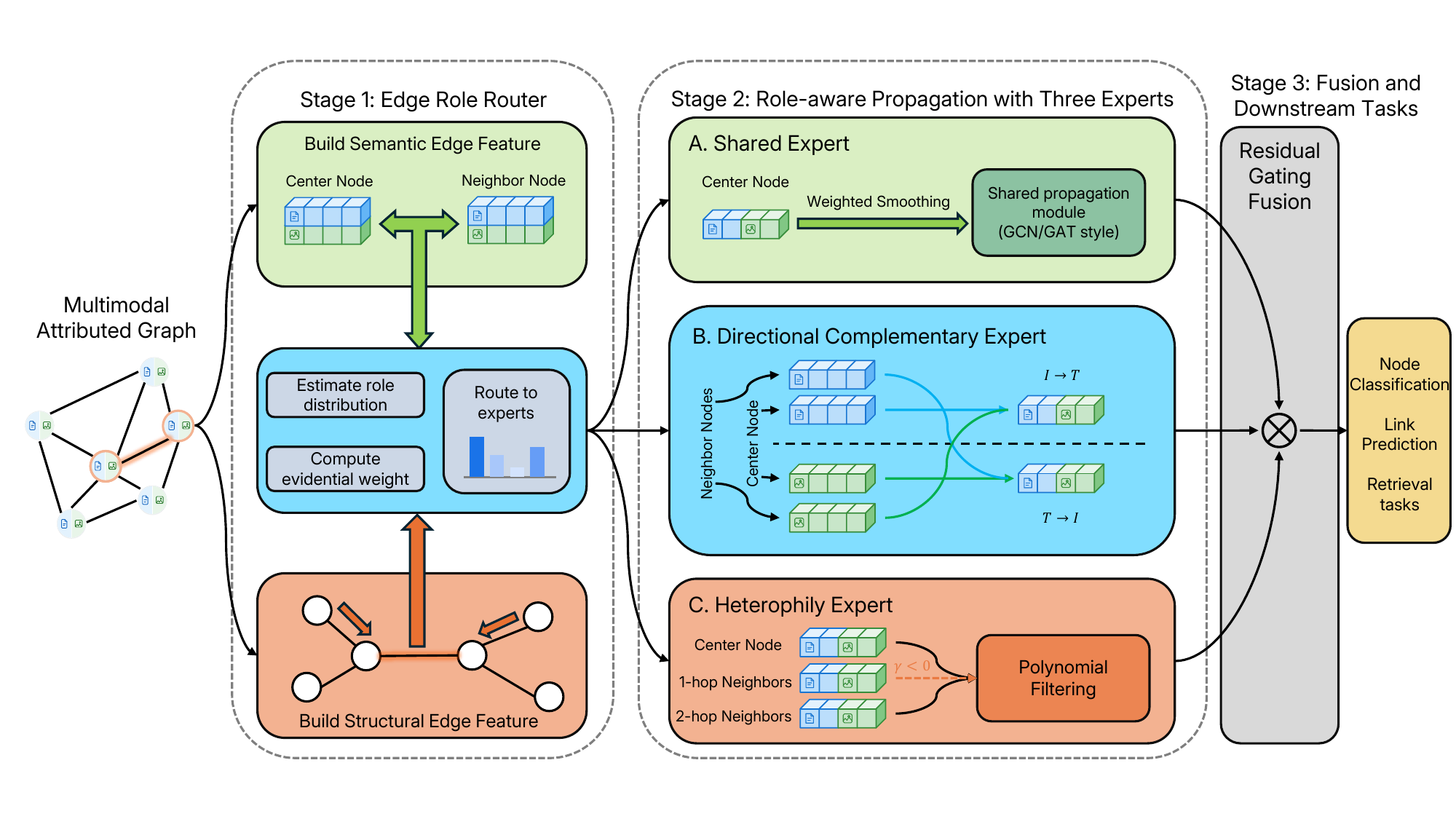}
    \caption{\textbf{Framework of RoleMAG.} RoleMAG performs role-aware multimodal propagation in three stages. First, an edge role router infers edge roles from semantic cues and uses structural features to calibrate routing strength. Second, the routed neighbors are sent to three specialized experts, including a shared expert for consistent smoothing, a directional complementary expert for cross-modal completion, and a heterophily expert for discrepant neighborhood interactions. Finally, the expert outputs are integrated by residual gating fusion and used for downstream tasks such as node classification, link prediction, and retrieval.}
    \label{fig:framework}
\end{figure*}

As shown in Fig.~\ref{fig:framework}, RoleMAG places the core of multimodal graph learning in the interaction and propagation stage~\cite{wan2026openmag}. Given the text and image representations of each node, the framework first constructs semantic and structural features for every edge. Semantic cues are used to estimate how that edge is distributed over three roles, namely \emph{shared}, \emph{complementary}, and \emph{heterophilous}, while structural cues are used to calibrate routing confidence. Shared neighbors are routed to a consistency-oriented propagation channel, complementary neighbors are sent to a directional cross-modal completion channel, and heterophilous neighbors are handled by a dedicated heterophily-aware propagator. The outputs of the three channels are finally fused through residual gating to obtain the task representation. This design follows recent discussions on structure-aware multimodal interaction~\cite{ning2025graph4mm,hong2026dip,jin2024instructg2i}, while also incorporating the basic insight from heterophily learning that heterophilous relations should not be directly mixed into shared smoothing~\cite{chien2021gprgnn,bo2021fagcn,li2022glognn}.

\subsection{Edge Role Estimation with Structural Confidence Calibration}

\noindent\textbf{Node representations.}
Let $\mathbf{h}_i^T,\mathbf{h}_i^I\in\mathbb{R}^{d}$ denote the text and image embeddings of node $i$, respectively. To provide a unified input for the later shared and heterophily-aware channels, a base node representation is first constructed as
\begin{equation}
\mathbf{h}_i = \phi \big([\mathbf{h}_i^T \Vert \mathbf{h}_i^I]\big),
\end{equation}
where $\phi(\cdot)$ is a linear projection or a lightweight MLP.

\noindent\textbf{Semantic and structural edge features.}
For each observed edge $(i,j)\in\mathcal{E}$, local semantic consistency is measured in both modalities:
\begin{equation}
s_{ij}^{T}=\cos(\mathrm{LN}(\mathbf{h}_i^T),\mathrm{LN}(\mathbf{h}_j^T)), \qquad
s_{ij}^{I}=\cos(\mathrm{LN}(\mathbf{h}_i^I),\mathrm{LN}(\mathbf{h}_j^I)).
\end{equation}
Here $\mathrm{LN}(\cdot)$ denotes layer normalization. These scores are used to form a three-dimensional semantic edge feature:
\begin{equation}
\mathbf{v}_{ij}^{\mathrm{sem}}
=
\big[
s_{ij}^{T},\;
s_{ij}^{I},\;
|s_{ij}^{T}-s_{ij}^{I}|
\big].
\end{equation}
The structural edge feature is defined by common local graph statistics:
\begin{equation}
\mathbf{v}_{ij}^{\mathrm{str}}
=
\big[
\tilde{A}_{ij},\;
\mathrm{CN}_{ij},\;
\mathrm{Jacc}_{ij},\;
\mathrm{AA}_{ij},\;
\mathrm{PA}_{ij},\;
\log(d_i+1),\;
\log(d_j+1)
\big],
\end{equation}
where $\mathrm{CN}$, $\mathrm{Jacc}$, $\mathrm{AA}$, and $\mathrm{PA}$ denote common neighbors, the Jaccard coefficient, Adamic--Adar, and preferential attachment, respectively.

\noindent\textbf{Factorized role variables.}
Rather than directly assigning the three roles through a black-box softmax classifier, RoleMAG first estimates two interpretable factors from the semantic edge feature:
\begin{equation}
\rho_{ij}^{T}=\sigma\!\big(g_T(\mathbf{v}_{ij}^{\mathrm{sem}})\big), \qquad
\rho_{ij}^{I}=\sigma\!\big(g_I(\mathbf{v}_{ij}^{\mathrm{sem}})\big),
\end{equation}
where $g_T(\cdot)$ and $g_I(\cdot)$ are lightweight MLPs. They characterize how semantically consistent and usable the edge is under the text and image modalities, respectively. The three-role distribution is then obtained analytically from these two factors:
\begin{equation}
\begin{aligned}
\pi_{ij}^{S} &= \rho_{ij}^{T}\rho_{ij}^{I}, \\
\pi_{ij}^{H} &= (1-\rho_{ij}^{T})(1-\rho_{ij}^{I}), \\
\pi_{ij}^{C} &= \rho_{ij}^{T}(1-\rho_{ij}^{I}) + (1-\rho_{ij}^{T})\rho_{ij}^{I}.
\end{aligned}
\label{eq:pi_factorized}
\end{equation}
Here $\pi_{ij}^{S}$ corresponds to shared support, $\pi_{ij}^{C}$ to complementary evidence, and $\pi_{ij}^{H}$ to heterophilous interaction.

\noindent\textbf{Evidence strength and effective routing weights.}
The role distribution alone is not sufficient to determine propagation strength. To address this, semantic strength and structural support are estimated separately as
\begin{equation}
s_{ij}^{\mathrm{sem}}=\mathrm{softplus}\!\big(g_{\mathrm{sem}}(\mathbf{v}_{ij}^{\mathrm{sem}})\big), \qquad
s_{ij}^{\mathrm{str}}=\mathrm{softplus}\!\big(g_{\mathrm{str}}(\mathbf{v}_{ij}^{\mathrm{str}})\big),
\end{equation}
and combined multiplicatively to produce the total edge-level evidence:
\begin{equation}
\beta_{ij}=s_{ij}^{\mathrm{sem}} \cdot s_{ij}^{\mathrm{str}}.
\label{eq:beta}
\end{equation}
This decomposition is consistent with the mean--precision separation used in evidential learning and prior networks~\cite{sensoy2018evidential,malinin2018prior}: the role distribution determines which type of edge it resembles, while $\beta_{ij}$ reflects how confident the model is in that judgment. The edge-level Dirichlet parameters are therefore written as
\begin{equation}
\mathbf{e}_{ij}=\beta_{ij}\boldsymbol{\pi}_{ij}, \qquad
\boldsymbol{\alpha}_{ij}=\mathbf{1}+\mathbf{e}_{ij},
\end{equation}
where $\boldsymbol{\pi}_{ij}=[\pi_{ij}^{S},\pi_{ij}^{C},\pi_{ij}^{H}]$. An effective confidence coefficient for propagation is then defined as
\begin{equation}
c_{ij}=\frac{\beta_{ij}}{3+\beta_{ij}},
\end{equation}
which yields the effective adjacency weights for the three channels:
\begin{equation}
\bar{A}_{ij}^{S}=c_{ij}\pi_{ij}^{S}, \qquad
\bar{A}_{ij}^{C}=c_{ij}\pi_{ij}^{C}, \qquad
\bar{A}_{ij}^{H}=c_{ij}\pi_{ij}^{H}.
\label{eq:effective_adj}
\end{equation}


\subsection{Shared Propagation via Role-weighted Neighborhood Smoothing}

\noindent\textbf{Shared expert.}
The shared channel models consistency propagation that benefits both modalities, and therefore adopts the simplest role-weighted smoothing scheme:
\begin{equation}
\mathbf{z}_i^{S}
=
\sum_{j\in\mathcal{N}(i)}
\bar{A}_{ij}^{S}\mathbf{W}_{S}\mathbf{h}_j.
\label{eq:shared_expert}
\end{equation}
This path preserves the low-pass behavior of standard message passing, but it is applied only on edges judged as \emph{shared} by the router. As a result, not all neighbors are indiscriminately mixed into the same smoothing process.

\subsection{Directional Complementary Completion via Query Bottleneck}

\noindent\textbf{Directional decomposition.}
The empirical study has shown that complementary relations not only exist, but are often directional. RoleMAG therefore separates the total complementary strength from its direction. Two unnormalized directional terms are first defined:
\begin{equation}
\tilde{d}_{ij}^{T\to I}=\rho_{ij}^{T}(1-\rho_{ij}^{I}), \qquad
\tilde{d}_{ij}^{I\to T}=(1-\rho_{ij}^{T})\rho_{ij}^{I},
\end{equation}
which are then normalized as
\begin{equation}
d_{ij}^{T\to I}
=
\frac{\tilde{d}_{ij}^{T\to I}}
{\tilde{d}_{ij}^{T\to I}+\tilde{d}_{ij}^{I\to T}+\epsilon},
\qquad
d_{ij}^{I\to T}
=
\frac{\tilde{d}_{ij}^{I\to T}}
{\tilde{d}_{ij}^{T\to I}+\tilde{d}_{ij}^{I\to T}+\epsilon}.
\label{eq:direction_norm}
\end{equation}
Accordingly, the direction-aware complementary weights are
\begin{equation}
\bar{A}_{ij}^{T\to I}=\bar{A}_{ij}^{C}d_{ij}^{T\to I}, \qquad
\bar{A}_{ij}^{I\to T}=\bar{A}_{ij}^{C}d_{ij}^{I\to T}.
\end{equation}

\noindent\textbf{Directional neighbor selection.}
To reduce the semantic noise introduced by redundant neighbors, only the strongest $K$ complementary neighbors are retained for each direction:
\begin{equation}
\mathcal{N}_{T\to I}(i)=\mathrm{TopK}_{j\in\mathcal{N}(i)}(\bar{A}_{ji}^{T\to I}), \qquad
\mathcal{N}_{I\to T}(i)=\mathrm{TopK}_{j\in\mathcal{N}(i)}(\bar{A}_{ji}^{I\to T}).
\label{eq:topk_dir}
\end{equation}

\noindent\textbf{Query-based complementary completion.}
The goal of this channel is not to perform another round of ordinary aggregation, but to extract from the opposite-modality neighborhood the completion evidence that is most useful for the current modality. A query-bottleneck design is adopted here, in line with Q-Former in BLIP-2 and subsequent graph-conditioned multimodal studies~\cite{li2023blip2,ning2025graph4mm,jin2024instructg2i}. Specifically, $R$ learnable query tokens are introduced for each direction:
\begin{equation}
\mathbf{q}_{i,r}^{T\to I}=\mathbf{q}_{r}^{T\to I}+\mathbf{W}_{QT}\mathbf{h}_{i}^{T}, \qquad
\mathbf{q}_{i,r}^{I\to T}=\mathbf{q}_{r}^{I\to T}+\mathbf{W}_{QI}\mathbf{h}_{i}^{I}.
\end{equation}
Here $T\!\to\!I$ means that text semantics are used to retrieve and complete image-side evidence, so the key/value pairs come from neighboring image representations; $I\!\to\!T$ is defined symmetrically:
\begin{equation}
\mathbf{k}_{j}^{I}=\mathbf{W}_{K}^{I}\mathbf{h}_{j}^{I}, \quad
\mathbf{v}_{j}^{I}=\mathbf{W}_{V}^{I}\mathbf{h}_{j}^{I},
\qquad
\mathbf{k}_{j}^{T}=\mathbf{W}_{K}^{T}\mathbf{h}_{j}^{T}, \quad
\mathbf{v}_{j}^{T}=\mathbf{W}_{V}^{T}\mathbf{h}_{j}^{T}.
\end{equation}

\noindent\textbf{Routing-aware attention bias.}
To let complementary routing directly affect the cross-modal retrieval process, the directional complementary weights are injected into the attention logits as a prior. This follows the same spirit as Graphormer, which encodes structural relations as attention bias~\cite{ying2021graphormer}. Taking $T\!\to\!I$ as an example,
\begin{equation}
\ell_{ij,r}^{T\to I}
=
\frac{(\mathbf{q}_{i,r}^{T\to I})^{\top}\mathbf{k}_{j}^{I}}{\sqrt{d}}
+
\lambda_{\mathrm{bias}}\log(\bar{A}_{ji}^{T\to I}+\epsilon),
\qquad j\in\mathcal{N}_{T\to I}(i).
\label{eq:qca_logit_t2i}
\end{equation}
After softmax normalization, the attention weights and query outputs are
\begin{equation}
a_{ij,r}^{T\to I}
=
\mathrm{softmax}_{j\in\mathcal{N}_{T\to I}(i)}
(\ell_{ij,r}^{T\to I}), \qquad
\mathbf{o}_{i,r}^{T\to I}
=
\sum_{j\in\mathcal{N}_{T\to I}(i)}
a_{ij,r}^{T\to I}\mathbf{v}_{j}^{I},
\end{equation}
and all query outputs are then pooled as
\begin{equation}
\mathbf{z}_{i}^{T\to I}
=
\mathrm{Pool}_{r=1}^{R}
(\mathbf{o}_{i,r}^{T\to I}).
\end{equation}
The $I\!\to\!T$ direction is fully symmetric. The final output of the complementary expert is
\begin{equation}
\mathbf{z}_{i}^{C}
=
\mathbf{W}_{C}
\big[
\mathbf{z}_{i}^{T\to I}
\Vert
\mathbf{z}_{i}^{I\to T}
\big].
\label{eq:comp_out}
\end{equation}

\subsection{Heterophily-aware Propagation via Signed Polynomial Filtering}

\noindent\textbf{Heterophily expert.}
For edges identified as heterophilous, they are no longer merged into shared smoothing. Instead, they are modeled by a dedicated signed polynomial filter. This design is consistent with the main idea behind GPR-GNN, FAGCN, and GloGNN: useful relations in heterophilous graphs often contain non-negligible high-frequency components, and propagation weights should not be restricted to nonnegative values~\cite{chien2021gprgnn,bo2021fagcn,li2022glognn}. Let $\tilde{\mathbf{A}}^{H}$ be the normalized heterophilous adjacency derived from $\bar{\mathbf{A}}^{H}$, then
\begin{equation}
\mathbf{Z}^{H}
=
\gamma_{0}\mathbf{H}
+
\gamma_{1}\tilde{\mathbf{A}}^{H}\mathbf{H}
+
\gamma_{2}(\tilde{\mathbf{A}}^{H})^{2}\mathbf{H},
\label{eq:polyh2}
\end{equation}
where $\mathbf{H}=[\mathbf{h}_{1},\ldots,\mathbf{h}_{N}]^{\top}$, and $\gamma_{0},\gamma_{1},\gamma_{2}$ are learnable coefficients that are allowed to take signed values. The $i$-th row is denoted by $\mathbf{z}_{i}^{H}$.

\subsection{Representation Fusion and Training Objective}

\noindent\textbf{Residual gating fusion.}
Since the three experts operate on different propagation regimes, the final representation is not formed by fixed summation. Instead, a lightweight gating network adaptively fuses them:
\begin{equation}
[g_{i}^{S},g_{i}^{C},g_{i}^{H}]
=
\mathrm{softmax}
\Big(
\mathrm{MLP}
\big(
[\mathbf{h}_{i}\Vert\mathbf{z}_{i}^{S}\Vert\mathbf{z}_{i}^{C}\Vert\mathbf{z}_{i}^{H}]
\big)
\Big),
\end{equation}
\begin{equation}
\mathbf{z}_{i}
=
\mathbf{h}_{i}
+
g_{i}^{S}\mathbf{z}_{i}^{S}
+
g_{i}^{C}\mathbf{z}_{i}^{C}
+
g_{i}^{H}\mathbf{z}_{i}^{H}.
\label{eq:fusion}
\end{equation}
The final task prediction is produced by a lightweight task head, whose supervision is denoted by $\mathcal{L}_{\mathrm{task}}$.

\noindent\textbf{Training objective.}
During training, auxiliary constraints are further imposed on the complementary expert and the router in addition to the main task objective. The overall loss is
\begin{equation}
\mathcal{L}
=
\mathcal{L}_{\mathrm{task}}
+
\lambda_{\mathrm{evi}}\mathcal{L}_{\mathrm{evi}}
+
\lambda_{\mathrm{qca}}\mathcal{L}_{\mathrm{qca}}
+
\lambda_{\mathrm{bal}}\mathcal{L}_{\mathrm{bal}}.
\label{eq:overall_loss}
\end{equation}

\noindent\textbf{Cross-modal completion alignment.}
To ensure that the complementary expert learns genuinely useful cross-modal completion rather than arbitrary attention reweighting, a contrastive alignment constraint is imposed on the completion outputs of both directions. Taking $T\!\to\!I$ as an example,
\begin{equation}
\mathbf{u}_{i}^{T\to I}=p_{I}(\mathbf{z}_{i}^{T\to I}), \qquad
\mathbf{v}_{i}^{I}=p_{I}(\mathbf{h}_{i}^{I}),
\end{equation}
\begin{equation}
\mathcal{L}_{\mathrm{qca}}^{T\to I}
=
-\frac{1}{|\Omega|}
\sum_{i\in\Omega}
\log
\frac{
\exp\big(\cos(\mathbf{u}_{i}^{T\to I},\mathbf{v}_{i}^{I})/\tau\big)
}{
\sum_{j\in\Omega}
\exp\big(\cos(\mathbf{u}_{i}^{T\to I},\mathbf{v}_{j}^{I})/\tau\big)
},
\end{equation}
where $\Omega$ is the set of nodes participating in the contrastive objective within the mini-batch. The $I\!\to\!T$ direction is defined in the same way, and the final loss is
\begin{equation}
\mathcal{L}_{\mathrm{qca}}
=
\mathcal{L}_{\mathrm{qca}}^{T\to I}
+
\mathcal{L}_{\mathrm{qca}}^{I\to T}.
\end{equation}

\noindent\textbf{Evidential regularization.}
To encourage the router to learn the evidence pattern that observed edges should be more certain while rewired edges should remain more uncertain, edge-level regularization is performed in the Dirichlet space. Let $\mathcal{E}_{\mathrm{in}}$ be the observed edges and $\mathcal{E}_{\mathrm{out}}$ the pseudo edges generated by random rewiring. Then
\begin{equation}
\begin{aligned}
\mathcal{L}_{\mathrm{evi}}
=
&\frac{1}{|\mathcal{E}_{\mathrm{in}}|}
\sum_{(i,j)\in\mathcal{E}_{\mathrm{in}}}
\mathrm{KL}
\big[
\mathrm{Dir}(\boldsymbol{\alpha}_{ij})
\;\|\;
\mathrm{Dir}(\hat{\boldsymbol{\alpha}}_{ij})
\big] \\
&+
\frac{1}{|\mathcal{E}_{\mathrm{out}}|}
\sum_{(i,j)\in\mathcal{E}_{\mathrm{out}}}
\mathrm{KL}
\big[
\mathrm{Dir}(\boldsymbol{\alpha}_{ij})
\;\|\;
\mathrm{Dir}(\mathbf{1})
\big],
\end{aligned}
\label{eq:evi_loss}
\end{equation}
where
\begin{equation}
\hat{\boldsymbol{\alpha}}_{ij}
=
\mathbf{1}
+
\beta_{\mathrm{in}}
\cdot
\mathrm{onehot}
\big(
\arg\max_{r\in\{S,C,H\}}\pi_{ij}^{r}
\big).
\end{equation}
This design follows the use of Dirichlet parameters in evidential learning and prior networks~\cite{sensoy2018evidential,malinin2018prior}, but shifts the target from node-class prediction to edge-role routing.

\noindent\textbf{Role balancing.}
Finally, to avoid the early-stage collapse of all edges into a single role, a lightweight batch-level balancing regularizer is introduced. Let the total soft assignment mass of the three roles within a mini-batch be
\begin{equation}
\mathrm{Imp}_{r}
=
\sum_{(i,j)\in\mathcal{E}_{b}}
\pi_{ij}^{r},
\qquad
r\in\{S,C,H\},
\end{equation}
then
\begin{equation}
\mathcal{L}_{\mathrm{bal}}
=
\mathrm{CV}
\big(
\{\mathrm{Imp}_{S},\mathrm{Imp}_{C},\mathrm{Imp}_{H}\}
\big)^{2},
\label{eq:balance}
\end{equation}
where $\mathrm{CV}(\cdot)$ denotes the coefficient of variation. This term only discourages extreme imbalance of role assignments at the batch level and does not alter the semantic judgment of any individual edge.

Overall, the training signals of RoleMAG are driven by three parts: $\mathcal{L}_{\mathrm{task}}$ optimizes the final representation for the downstream task, $\mathcal{L}_{\mathrm{qca}}$ constrains the complementary expert to learn meaningful cross-modal completion, and $\mathcal{L}_{\mathrm{evi}}$ together with $\mathcal{L}_{\mathrm{bal}}$ stabilizes the evidence-learning process of the edge router.

\begin{table*}[t]
\centering
\caption{Main results on graph-centric tasks. All metrics are reported as percentages. The best, second best, and third best results are highlighted in dark red, royal blue, and orange, respectively.}
\label{tab:main_results_graph}
\resizebox{\textwidth}{!}{
\begin{tabular}{lcccccc}
\toprule
\multirow{2}{*}{Model} & \multicolumn{2}{c}{Toys (Node Classification)} & \multicolumn{2}{c}{RedditS (Node Classification)} & \multicolumn{2}{c}{Bili\_Dance (Link Prediction)} \\
\cmidrule(lr){2-3} \cmidrule(lr){4-5} \cmidrule(lr){6-7}
& ACC & F1 & ACC & F1 & MRR & Hits@3 \\
\midrule
GCN 
& 77.24$_{\pm 0.31}$ & 73.18$_{\pm 0.42}$ 
& 91.48$_{\pm 0.24}$ & 84.68$_{\pm 0.38}$ 
& 14.25$_{\pm 0.41}$ & 15.23$_{\pm 0.57}$ \\

GAT 
& 77.29$_{\pm 0.34}$ & 72.92$_{\pm 0.49}$ 
& 91.45$_{\pm 0.27}$ & 85.65$_{\pm 0.41}$ 
& \textcolor{orange}{\textbf{24.19$_{\pm 0.88}$}} & \textcolor{orange}{\textbf{25.42$_{\pm 1.06}$}} \\

MMGCN 
& 77.60$_{\pm 0.29}$ & \textcolor{orange}{\textbf{75.03$_{\pm 0.45}$}} 
& 90.44$_{\pm 0.36}$ & 82.85$_{\pm 0.56}$ 
& 8.64$_{\pm 0.33}$ & 8.55$_{\pm 0.39}$ \\

MGAT 
& 78.12$_{\pm 0.42}$ & 74.21$_{\pm 0.61}$ 
& 91.73$_{\pm 0.31}$ & 85.12$_{\pm 0.48}$ 
& 16.84$_{\pm 0.51}$ & 18.92$_{\pm 0.72}$ \\

LGMRec 
& \textcolor{orange}{\textbf{78.81$_{\pm 0.73}$}} & 72.31$_{\pm 1.14}$ 
& \textcolor{royalblue}{\textbf{92.36$_{\pm 0.19}$}} & \textcolor{orange}{\textbf{86.10$_{\pm 0.32}$}} 
& 11.46$_{\pm 0.37}$ & 11.78$_{\pm 0.44}$ \\

DGF 
& 77.68$_{\pm 0.38}$ & 71.46$_{\pm 0.58}$ 
& 91.79$_{\pm 0.26}$ & 83.03$_{\pm 0.47}$ 
& 7.29$_{\pm 0.28}$ & 7.29$_{\pm 0.31}$ \\

DMGC 
& 67.89$_{\pm 0.55}$ & 58.42$_{\pm 0.83}$ 
& 76.57$_{\pm 0.62}$ & 65.08$_{\pm 0.95}$ 
& 13.14$_{\pm 0.45}$ & 15.01$_{\pm 0.53}$ \\

NTSFormer 
& 78.54$_{\pm 0.33}$ & 74.88$_{\pm 0.52}$ 
& 92.21$_{\pm 0.24}$ & 85.74$_{\pm 0.40}$ 
& 20.31$_{\pm 0.64}$ & 24.66$_{\pm 0.79}$ \\

Graph4MM 
& \textcolor{darkred}{\textbf{78.91$_{\pm 0.30}$}} & \textcolor{darkred}{\textbf{75.62$_{\pm 0.48}$}} 
& \textcolor{orange}{\textbf{92.34$_{\pm 0.22}$}} & \textcolor{royalblue}{\textbf{86.31$_{\pm 0.37}$}} 
& \textcolor{royalblue}{\textbf{28.42$_{\pm 0.71}$}} & \textcolor{royalblue}{\textbf{34.85$_{\pm 0.95}$}} \\

RoleMAG 
& \textcolor{royalblue}{\textbf{78.83$_{\pm 0.27}$}} & \textcolor{royalblue}{\textbf{75.49$_{\pm 0.41}$}} 
& \textcolor{darkred}{\textbf{92.57$_{\pm 0.18}$}} & \textcolor{darkred}{\textbf{87.04$_{\pm 0.31}$}} 
& \textcolor{darkred}{\textbf{30.28$_{\pm 0.82}$}} & \textcolor{darkred}{\textbf{35.91$_{\pm 1.02}$}} \\
\bottomrule
\end{tabular}
}
\end{table*}

\begin{table}[t]
\centering
\caption{Ablation study of RoleMAG on node classification. All metrics are reported as percentages.}
\label{tab:ablation}
\resizebox{\columnwidth}{!}{
\begin{tabular}{lcc}
\toprule
Variant & Toys ACC & RedditS ACC \\
\midrule
Shared-only & 75.23 & 89.73 \\
w/o Role Routing & 77.23 & 90.85 \\
w/o Complementary Expert & 77.85 & 91.44 \\
w/o Complementary Direction & 77.93 & 91.62 \\
w/o Heterophily Expert & 76.97 & 90.44 \\
\textbf{Full RoleMAG} & \textcolor{darkred}{\textbf{78.83}} & \textcolor{darkred}{\textbf{92.57}} \\
\bottomrule
\end{tabular}
}
\end{table}

\section{Experiments}

In this section, we first describe the experimental setup, with full implementation details, hyperparameter ranges, and additional results deferred to the supplementary material. We then conduct empirical evaluations to answer the following research questions: \textbf{Q1}: Can RoleMAG outperform representative graph and multimodal graph baselines on graph-centric tasks? \textbf{Q2}: What are the individual contributions of role routing, the complementary expert, and the heterophily expert? \textbf{Q3}: How robust is RoleMAG under structural perturbation? \textbf{Q4}: What is the efficiency--effectiveness trade-off of RoleMAG?

\subsection{Experimental Setup}

\noindent\textbf{Datasets.}
We conduct the main-paper evaluation on three representative graph-centric benchmarks from OpenMAG~\cite{wan2026openmag}, following its unified MAG evaluation protocol and the experimental organization adopted in TMTE~\cite{zhu2026tmte}. Specifically, \textbf{Toys} and \textbf{RedditS} are used for node classification, while \textbf{Bili\_Dance} is used for link prediction. These datasets cover recommendation, social media, and video recommendation scenarios, respectively, and expose different types of neighborhood interactions. Due to space limitations, detailed statistics and preprocessing pipelines are provided in the supplementary material.

\noindent\textbf{Baselines.}
To ensure representative rather than exhaustive comparison, we select strong baselines from three groups. 
(1) \textbf{Unimodal GNN backbones}: GCN~\cite{kipf2017semi} and GAT~\cite{velickovic2018graph}. 
(2) \textbf{Classical multimodal graph models}: MMGCN~\cite{wei2019mmgcn}, MGAT~\cite{tao2020mgat}, and LGMRec~\cite{guo2024lgmrec}. 
(3) \textbf{Recent graph-enhanced or Transformer-enhanced MAG models}: DGF, DMGC, NTSFormer, and Graph4MM~\cite{ning2025graph4mm}. 
This subset is sufficient to cover shared propagation, multimodal interaction, structural denoising, and recent query-based fusion designs, while keeping the main table readable.

\noindent\textbf{Downstream Tasks and Metrics.}
We focus on two graph-centric tasks in the main paper: \textbf{node classification} and \textbf{link prediction}. For node classification, we report \textbf{ACC} and \textbf{F1}; for link prediction, we report \textbf{MRR} and \textbf{Hits@3}. Following the presentation style of TMTE~\cite{zhu2026tmte}, all metrics are reported in percentage form. Results in the main table are averaged over multiple runs and reported as mean$\pm$std when available.

\noindent\textbf{Implementation Note.}
The experiments are conducted under a unified benchmark setting, and all compared methods share the same data split and evaluation protocol. In the main paper, we concentrate on the results that most directly reflect the effect of propagation design. Additional datasets and more extensive comparisons can be moved to the supplementary material if needed.

\subsection{Overall Performance}

To answer \textbf{Q1}, we first compare RoleMAG with representative baselines on graph-centric tasks. The results are reported in Table~\ref{tab:main_results_graph}.

\textbf{Graph-centric tasks.}
As shown in Table~\ref{tab:main_results_graph}, RoleMAG achieves the best overall performance on \textbf{RedditS} and \textbf{Bili\_Dance}, and remains highly competitive on \textbf{Toys}. On RedditS, RoleMAG reaches \textcolor{darkred}{\textbf{92.57}} ACC and \textcolor{darkred}{\textbf{87.04}} F1, surpassing the strongest baseline by \textbf{+0.21} ACC and \textbf{+0.73} F1. On Bili\_Dance, RoleMAG further obtains \textcolor{darkred}{\textbf{30.28}} MRR and \textcolor{darkred}{\textbf{35.91}} Hits@3, improving over Graph4MM by \textbf{+1.86} MRR and \textbf{+1.06} Hits@3. These gains are not large in an absolute sense, but they are stable across both metrics, which is more important for a propagation-oriented method.

\textbf{Comparison with strong multimodal baselines.}
A more careful observation is that the strongest competitors differ across datasets. On Toys, Graph4MM achieves the best ACC and F1, while RoleMAG ranks second and stays very close, with only \textbf{0.08} ACC and \textbf{0.13} F1 behind. This gap is small enough to suggest that RoleMAG does not rely on a narrow dataset-specific advantage. In contrast, once the task becomes more sensitive to mixed neighborhood signals, as in RedditS and Bili\_Dance, the benefit of role-aware propagation becomes clearer. In these settings, separating shared, complementary, and heterophilous neighbors appears more useful than relying on a single shared propagation path.

\textbf{Discussion.}
Overall, the main result does not suggest that RoleMAG dominates every setting by a large margin. The picture is more precise than that. RoleMAG is strongest when neighborhood interactions are harder to use with a single propagation rule, and it remains competitive when a strong query-based baseline already performs well. This behavior is consistent with the design objective of RoleMAG: the method is intended to improve \emph{how} neighbors participate in propagation, rather than to replace all existing multimodal fusion mechanisms.
\begin{figure}[t]
\centering
\begin{minipage}{0.48\columnwidth}
    \centering
    \includegraphics[width=\linewidth]{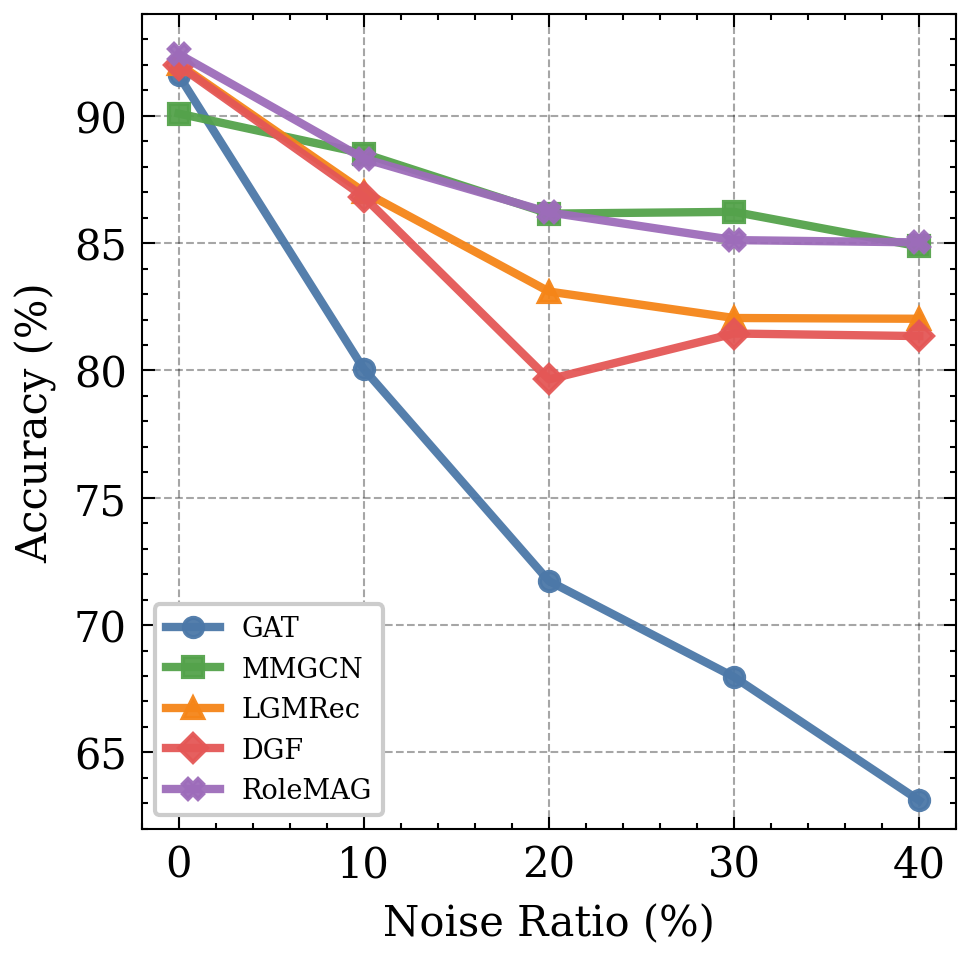}
    \vspace{2pt}
    \centerline{\small (a) Robustness under structural perturbation}
\end{minipage}\hfill
\begin{minipage}{0.48\columnwidth}
    \centering
    \includegraphics[width=\linewidth]{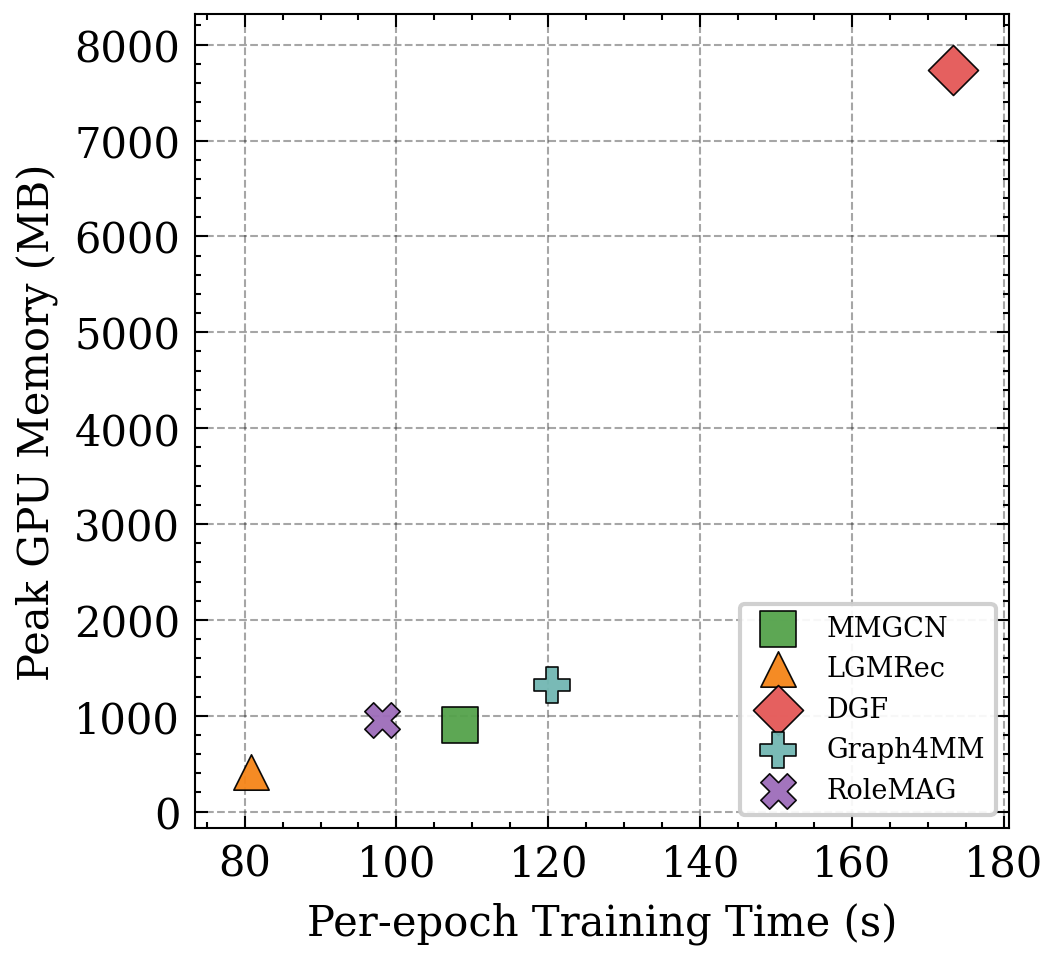}
    \vspace{2pt}
    \centerline{\small (b) Efficiency comparison}
\end{minipage}
\caption{Robustness and efficiency analysis.}
\label{fig:robust_eff}
\end{figure}

\subsection{Ablation Study}

To answer \textbf{Q2}, we conduct an ablation study to assess the contribution of each component in RoleMAG. Since the three propagation channels are coupled through the same routing distribution, the most informative way is not to remove arbitrary operations in isolation, but to define several meaningful variants that break one design choice at a time.

\textbf{All components contribute to the final performance.}
Table~\ref{tab:ablation} shows a clear and consistent pattern: the full model performs best on both Toys and RedditS. Once any key component is removed, performance drops. This is the most basic but also the most important observation, because it indicates that the gains of RoleMAG do not come from one accidental trick.

\textbf{Shared propagation alone is insufficient.}
The \emph{Shared-only} variant causes the largest overall degradation, dropping from \textbf{78.83} to \textbf{75.23} on Toys and from \textbf{92.57} to \textbf{89.73} on RedditS. This result directly supports the main motivation of the paper: forcing all neighbors into one common propagation path is overly restrictive. It mixes together signals that should not be treated in the same way.

\textbf{The heterophily expert is especially important.}
Among the single-module removals, \emph{w/o Heterophily Expert} yields the most pronounced drop, especially on RedditS, where the accuracy falls by \textbf{2.13} points. The same trend also appears on Toys. This suggests that keeping heterophilous neighbors out of shared smoothing is not a marginal design choice. It is one of the main reasons why RoleMAG remains effective when neighborhood patterns are mixed.

\textbf{The complementary channel is not decorative.}
Removing the complementary expert or discarding directional modeling also leads to stable degradation. On Toys, \emph{w/o Complementary Expert} and \emph{w/o Complementary Direction} reduce accuracy to \textbf{77.85} and \textbf{77.93}, respectively. On RedditS, the corresponding numbers are \textbf{91.44} and \textbf{91.62}. The drop is smaller than that caused by removing the heterophily expert, yet it remains consistent across both datasets. This is important: the complementary channel is not simply an auxiliary branch added for multimodal flavor. It contributes measurable gains once the model is asked to distinguish how neighbors should help each modality.

\subsection{Robustness Analysis}

To answer \textbf{Q3}, we investigate the robustness of RoleMAG under structural perturbation, following the topology-noise evaluation perspective emphasized in OpenMAG and TMTE~\cite{wan2026openmag,zhu2026tmte}. We gradually inject edge noise into the RedditS graph and compare different models under increasing perturbation ratios.

\textbf{RoleMAG remains strong under noisy neighborhoods.}
As illustrated in Fig.~\ref{fig:robust_eff}(a), all methods suffer performance degradation as the noise ratio increases, but the degradation patterns differ substantially. GAT drops rapidly from \textbf{91.60} to \textbf{63.11}, showing that a standard attention-based aggregator is highly sensitive to structural corruption. MMGCN and LGMRec are more stable, yet their performance also declines steadily once the perturbation becomes stronger.

\textbf{The advantage of RoleMAG becomes clearer at moderate-to-high noise levels.}
A more informative pattern is that RoleMAG is already the best model at the clean setting, stays near the top at \textbf{10\%} noise, and becomes the best model again from \textbf{20\%} to \textbf{40\%} noise. In particular, it reaches \textbf{86.21}, \textbf{85.12}, and \textbf{85.03} ACC at \textbf{20\%}, \textbf{30\%}, and \textbf{40\%} noise, respectively, all higher than the compared baselines. This result is consistent with the design of RoleMAG: once the graph becomes less reliable, it is increasingly important to avoid treating all neighbors as equally trustworthy participants in message passing.

\subsection{Efficiency Analysis}

To answer \textbf{Q4}, we evaluate the computational overhead of RoleMAG using the quantities that are consistently available in the current logging pipeline, namely per-epoch training time and peak GPU memory. The comparison is summarized in Fig.~\ref{fig:robust_eff}(b).

\textbf{RoleMAG is not the cheapest model, but its overhead remains moderate.}
Compared with lightweight baselines such as LGMRec, RoleMAG introduces additional cost. Its per-epoch training time is \textbf{98.12}s, compared with \textbf{80.85}s for LGMRec, and its peak memory is \textbf{951}MB, compared with \textbf{412}MB. This extra overhead is expected, since RoleMAG explicitly models three propagation roles instead of relying on a single shared path.

\textbf{RoleMAG remains clearly cheaper than several strong competitors.}
More importantly, RoleMAG is still more efficient than heavier baselines that are closer in modeling strength. Relative to Graph4MM, RoleMAG reduces the per-epoch training time from \textbf{120.45}s to \textbf{98.12}s and the peak memory from \textbf{1322}MB to \textbf{951}MB, while achieving stronger overall graph-task performance. The gap becomes even larger when compared with DGF, whose peak memory reaches \textbf{7734}MB. Therefore, the extra cost of RoleMAG is better described as \emph{moderate} rather than excessive.

\textbf{Discussion.}
Taken together, these results indicate that RoleMAG achieves a favorable trade-off between effectiveness and efficiency. It is not designed to minimize overhead at all costs. Instead, it uses a moderate amount of additional computation to obtain more reliable propagation behavior, especially on datasets where mixed neighborhood roles matter most.
\section{Conclusion}

In this paper, we revisit a basic assumption in multimodal graph learning: that the same neighbors can be propagated in largely the same way across modalities. Our empirical study shows that this view is often too coarse, since neighborhood utility can be modality-dependent, complementary signals may be directional, and heterophilous relations should not be directly mixed into shared smoothing. Motivated by these observations, RoleMAG is designed as a role-aware propagation framework that estimates the distribution of shared, complementary, and heterophilous roles at the edge level and routes them through separate channels. In particular, the complementary channel supports directional cross-modal completion, while the heterophily-aware channel preserves informative but dissimilar relations outside the shared propagation path. Experiments on benchmark Multimodal attributed graphs demonstrate that this design consistently improves graph-centric performance, remains robust under structural perturbation, and achieves a favorable effectiveness--efficiency trade-off. Overall, these results suggest that modeling how neighbors should participate in multimodal propagation can be as important as modeling the modalities themselves.

\bibliographystyle{ACM-Reference-Format}
\bibliography{appendix_references}
\clearpage
\appendix
\section{Related Work}
\label{app:related_work}

\noindent\textbf{Multimodal graph learning and MAG benchmarks.}
Multimodal graph learning aims to jointly model graph topology and multimodal node attributes, and its problem space and methodological landscape have been systematically reviewed in prior surveys\cite{ektefaie2023multimodal}. In earlier application-driven studies, multimodal signals were mainly introduced as auxiliary cues to improve task-specific graph learning. For instance, MMGCN\cite{wei2019mmgcn} performs parallel propagation over different modalities. MGAT\cite{tao2020mgat} and DualGNN\cite{wang2023dualgnn} further distinguish user preferences across modalities. LATTICE\cite{zhang2021lattice} and FREEDOM\cite{zhou2023freedom} shift the focus toward mining or preserving latent item--item structures from multimodal content. MMSSL\cite{wei2023mmssl} and BM3\cite{zhou2023bm3} strengthen cross-modal alignment and representation robustness through self-supervised objectives, while LGMRec\cite{guo2024lgmrec} combines local and global graph signals to improve modeling under sparse interactions. Meanwhile, the evaluation ecosystem has evolved from MM-GRAPH\cite{zhu2025mmgraph} to OpenMAG\cite{wan2026openmag}, providing a more unified context for datasets, models, and tasks in MAG research. These studies have substantially advanced the field. Yet most of them still center on modality-level preference modeling, graph structure refinement, or self-supervised training, and only rarely ask a more direct question in general MAGs: what role should different neighbors play during propagation?

\noindent\textbf{Structure-aware multimodal interaction and routing.}
As the field moves from asking how multimodal features should be used to asking how structure should enter multimodal interaction, increasing attention has been paid to the propagation stage itself. Graph4MM\cite{ning2025graph4mm} argues that earlier methods often treat the graph as an independent modality and do not distinguish multi-hop neighborhoods, and therefore injects structural context into foundation multimodal models through Hop-Diffused Attention and MM-QFormer. TMTE\cite{zhu2026tmte} studies task-aware topology and jointly updates modality representations and graph structure. DiP\cite{hong2026dip} further introduces pseudo-nodes and dynamic information paths to support more flexible intra-modal propagation and inter-modal aggregation. Mario\cite{sun2026mario}, from a large-model perspective, emphasizes weak cross-modal consistency and heterogeneous modality preferences, and designs a modality-adaptive router for this setting. In parallel with these graph-learning studies, BLIP-2\cite{li2023blip2} shows that a query-based bottleneck, instantiated as Q-Former, is an effective bridge for cross-modal interaction. InstructG2I\cite{jin2024instructg2i} further adopts a graph Q-Former to encode graph context for graph-to-image generation in MMAGs. Taken together, these studies suggest that information interaction in multimodal graphs should no longer be viewed as a uniform, static, and modality-agnostic process. Still, their main focus remains on structure injection, path scheduling, topology co-evolution, or node-level modality preference. A finer-grained issue remains largely open: whether the same neighbor should provide \emph{shared support}, \emph{complementary completion}, or \emph{heterophilous interaction} for different modalities.

\noindent\textbf{Heterophily-aware propagation and graph filtering.}
Another line of work closely related to this paper comes from heterophily-aware graph learning. The central observation is that connected nodes are not always semantically similar. As a result, relying only on low-pass smoothing can weaken discriminability and may even amplify misleading propagation. GPR-GNN\cite{chien2021gprgnn} uses learnable generalized PageRank weights to adapt to both homophilous and heterophilous graphs. FAGCN\cite{bo2021fagcn} explicitly combines low- and high-frequency components, showing that high-frequency signals in heterophilous settings are not merely noise. GloGNN\cite{li2022glognn} further highlights the role of global node correlations, suggesting that local neighborhoods alone are often insufficient for recovering useful same-class information under heterophily. For the present work, these studies provide an important lesson: heterophilous relations should not be directly mixed into shared smoothing, but should instead be handled by a dedicated propagation mechanism. However, existing heterophily-aware methods are mostly developed for unimodal graph learning. They do not consider the MAG setting, where cross-modal complementarity and heterophilous relations may coexist and interact. RoleMAG starts from this gap. In multimodal graphs, the key question is not only whether to preserve low-frequency or high-frequency signals, but also how to distinguish the role that each neighbor plays in cross-modal propagation and route it accordingly.

\section{Method Specification}
\label{app:method_spec}

\zh{
\begin{table*}[t]
\centering
\small
\setlength{\tabcolsep}{5pt}
\renewcommand{\arraystretch}{1.12}
\begin{tabular}{p{0.22\textwidth} p{0.49\textwidth} p{0.21\textwidth}}
\hline
\textbf{Symbol} & \textbf{Meaning} & \textbf{Shape / Range} \\
\hline
\multicolumn{3}{l}{\textit{Input graph and node representations}} \\
$\mathcal{G}=(\mathcal{V},\mathcal{E},\{\mathbf{X}^{(m)}\}_{m\in\mathcal{M}})$
& Multimodal attributed graph
& -- \\
$\mathcal{M}=\{T,I\}$
& Set of modality indices, representing text and image, respectively
& $|\mathcal{M}|=2$ \\
$\mathbf{x}_i^{(m)},\mathbf{X}^{(m)}$
& Modal input features of node $v_i$ and the modal feature matrix of the entire graph
& $\mathbb{R}^{d_m}$, $\mathbb{R}^{N\times d_m}$ \\
$\mathbf{A},\mathbf{D},\tilde{\mathbf{A}}$
& Adjacency matrix, degree matrix, and symmetric normalized adjacency matrix
& $\mathbb{R}^{N\times N}$ \\
$\mathcal{N}(i)$
& First-order neighborhood of node $v_i$
& $\subseteq \mathcal{V}$ \\
$\mathbf{h}_i^{T},\mathbf{h}_i^{I}$
& Representation of node $v_i$ in text / image modality
& $\mathbb{R}^{d}$ \\
$\mathbf{h}_i,\mathbf{H}$
& Unified input representation and its matrix form
& $\mathbb{R}^{d}$, $\mathbb{R}^{N\times d}$ \\
\hline
\multicolumn{3}{l}{\textit{Router variables}} \\
$\mathbf{v}_{ij}^{\mathrm{sem}},\mathbf{v}_{ij}^{\mathrm{str}}$
& Semantic and structural edge features for edge $(i,j)$
& $\mathbb{R}^{d_{\mathrm{sem}}}$, $\mathbb{R}^{d_{\mathrm{str}}}$ \\
$\rho_{ij}^{T},\rho_{ij}^{I}$
& Modality availability factors for text / image
& $(0,1)$ \\
$\pi_{ij}^{S},\pi_{ij}^{C},\pi_{ij}^{H}$
& Role distribution of shared / complementary / heterophilous
& $[0,1]$, and sum to $1$ \\
$\beta_{ij},\boldsymbol{\alpha}_{ij},c_{ij}$
& Evidence strength, Dirichlet parameters, and effective confidence coefficient
& See \cref{app:edge_role_router} \\
$\bar{A}_{ij}^{S},\bar{A}_{ij}^{C},\bar{A}_{ij}^{H}$
& Effective edge weights of the three propagation channels
& $\mathbb{R}_{\ge 0}$ \\
\hline
\multicolumn{3}{l}{\textit{Expert and fusion variables}} \\
$d_{ij}^{T\rightarrow I},d_{ij}^{I\rightarrow T}$
& Directional decomposition weights in the complementary channel
& $[0,1]$ \\
$K,Q$
& Number of Top-$K$ neighbors and queries kept in each direction
& $\mathbb{N}_{+}$ \\
$R$
& Order of the signed polynomial filter
& $\mathbb{N}_{+}$ \\
$\lambda_{\mathrm{bias}}$
& Scaling coefficient of the routing-aware attention bias
& $\mathbb{R}_{\ge 0}$ \\
$\mathbf{z}_{i}^{T\rightarrow I},\mathbf{z}_{i}^{I\rightarrow T}$
& Complementary completion representations for the two directions
& $\mathbb{R}^{d}$ \\
$\mathbf{z}_i^{S},\mathbf{z}_i^{C},\mathbf{z}_i^{H}$
& Outputs of the three experts
& $\mathbb{R}^{d}$ \\
$\mathbf{z}_i,\mathbf{Z}$
& Final node representation after fusion and its matrix form
& $\mathbb{R}^{d}$, $\mathbb{R}^{N\times d}$ \\
\hline
\multicolumn{3}{l}{\textit{Objective-related variables}} \\
$\psi_{\mathrm{task}}(\cdot)$
& Downstream task head
& Task-dependent \\
$\mathcal{V}_L$
& Set of supervised nodes in the node classification task
& $\subseteq \mathcal{V}$ \\
$\Omega$
& Set of nodes participating in the contrastive objective in the current mini-batch
& $\subseteq \mathcal{V}$ \\
$\tau$
& Temperature parameter in the contrastive objective
& $\mathbb{R}_{>0}$ \\
$p_T(\cdot),p_I(\cdot)$
& Alignment space projection heads
& Task-specific mapping \\
$\mathcal{E}_{\mathrm{in}},\mathcal{E}_{\mathrm{out}}$
& Set of observed edges and pseudo edges
& $\subseteq \mathcal{V}\times\mathcal{V}$ \\
$\mathcal{E}_{b}$
& Set of edges in the current mini-batch
& $\subseteq \mathcal{E}$ \\
$\eta_{\mathrm{KL}}$
& Balancing coefficient for the KL term in evidential regularization
& $\mathbb{R}_{\ge 0}$ \\
$\mathcal{L}_{\mathrm{task}},\mathcal{L}_{\mathrm{qca}},\mathcal{L}_{\mathrm{evi}},\mathcal{L}_{\mathrm{bal}}$
& Primary task loss, completion constraint, evidential regularization, and role balancing
& Scalar \\
$\lambda_{\mathrm{qca}},\lambda_{\mathrm{evi}},\lambda_{\mathrm{bal}}$
& Balancing coefficients for each auxiliary loss term
& $\mathbb{R}_{\ge 0}$ \\
\hline
\end{tabular}
\caption{\textbf{Notation used in Appendix B.} The notations above are used consistently throughout this appendix. For detailed definitions of each variable, please refer to \cref{app:overview_notation,app:edge_role_router,app:shared_expert,app:complementary_expert,app:heterophily_expert,app:fusion_objectives}.}
\label{tab:rolemag_notation}
\end{table*}

\subsection{Overview and Notation Alignment}
\label{app:overview_notation}

This section supplements the method definition and the three-channel propagation pipeline presented in the main paper. Unless otherwise specified, the description below follows the same design as the main paper and uses the same notation to specify module interfaces, variables, and training objectives.

Let the multimodal attributed graph be denoted by
\[
\mathcal{G}=(\mathcal{V},\mathcal{E},\{\mathbf{X}^{(m)}\}_{m\in\mathcal{M}}),
\]
where $\mathcal{V}$ is the node set, $\mathcal{E}$ is the edge set, $N=|\mathcal{V}|$, and the modality index set is $\mathcal{M}=\{T,I\}$, corresponding to text and image, respectively. For any node $v_i\in\mathcal{V}$, its input feature under modality $m$ is denoted by $\mathbf{x}_i^{(m)}\in\mathbb{R}^{d_m}$, and the feature matrix collecting all node features under that modality is denoted by $\mathbf{X}^{(m)}\in\mathbb{R}^{N\times d_m}$. The adjacency matrix of the graph is written as $\mathbf{A}\in\mathbb{R}^{N\times N}$, the degree matrix as $\mathbf{D}$, and the corresponding symmetrically normalized adjacency matrix as
\[
\tilde{\mathbf{A}}=\mathbf{D}^{-\frac12}\mathbf{A}\mathbf{D}^{-\frac12}.
\]
The one-hop neighborhood of node $v_i$ is denoted by
\[
\mathcal{N}(i)=\{j\mid (i,j)\in\mathcal{E}\}.
\]

For each node, the text and image inputs are passed through modality encoders and projection layers to obtain modality-specific representations in a shared dimensionality:
\[
\mathbf{h}_i^{T}\in\mathbb{R}^{d},\qquad
\mathbf{h}_i^{I}\in\mathbb{R}^{d}.
\]
Consistent with the main paper, the unified input representation is obtained by concatenating the two modality representations and applying a learnable mapping $\phi(\cdot)$:
\begin{equation}
\mathbf{h}_i
=
\phi\!\big([\mathbf{h}_i^{T}\Vert \mathbf{h}_i^{I}]\big)
\in\mathbb{R}^{d}.
\label{eq:app_unified_input}
\end{equation}
Denote
\[
\mathbf{H}^{T}=[\mathbf{h}_1^{T};\dots;\mathbf{h}_N^{T}],\qquad
\mathbf{H}^{I}=[\mathbf{h}_1^{I};\dots;\mathbf{h}_N^{I}],\qquad
\mathbf{H}=[\mathbf{h}_1;\dots;\mathbf{h}_N].
\]
Here $\mathbf{H}$ serves as the shared input to the shared expert, the heterophily expert, and the final residual path. In contrast, $\mathbf{H}^{T}$ and $\mathbf{H}^{I}$ preserve modality-specific information and are used in the directional cross-modal interaction of the complementary expert. Details of the modality encoders, input construction, and experimental configurations are provided in Appendix D.

For any observed edge $(i,j)\in\mathcal{E}$, the router constructs a semantic edge feature and a structural edge feature, denoted by
\[
\mathbf{v}_{ij}^{\mathrm{sem}}\in\mathbb{R}^{d_{\mathrm{sem}}},\qquad
\mathbf{v}_{ij}^{\mathrm{str}}\in\mathbb{R}^{d_{\mathrm{str}}}.
\]
Based on these two types of edge features, the router estimates modality usability factors under the text and image modalities:
\[
\rho_{ij}^{T}\in(0,1),\qquad
\rho_{ij}^{I}\in(0,1),
\]
and derives a three-role distribution
\[
\boldsymbol{\pi}_{ij}
=
[\pi_{ij}^{S},\pi_{ij}^{C},\pi_{ij}^{H}],
\qquad
\pi_{ij}^{S}+\pi_{ij}^{C}+\pi_{ij}^{H}=1,
\]
where $S$, $C$, and $H$ denote shared, complementary, and heterophilous, respectively. In addition, the router outputs an evidence strength $\beta_{ij}$, Dirichlet parameters $\boldsymbol{\alpha}_{ij}$, and an effective confidence coefficient $c_{ij}$ for each edge, together with the effective edge weights of the three propagation channels:
\[
\bar{A}_{ij}^{S},\qquad
\bar{A}_{ij}^{C},\qquad
\bar{A}_{ij}^{H}.
\]

On this basis, the propagation pipeline of RoleMAG consists of three experts. The shared expert takes $\mathbf{H}$ and $\bar{\mathbf{A}}^{S}$ as input and outputs
\[
\mathbf{z}_i^{S}\in\mathbb{R}^{d},
\]
which models support signals over shared and consistent neighborhoods. The complementary expert takes $\mathbf{H}^{T}$, $\mathbf{H}^{I}$, and $\bar{\mathbf{A}}^{C}$ as input, and further decomposes the complementary channel into two directions:
\[
d_{ij}^{T\rightarrow I},\qquad
d_{ij}^{I\rightarrow T},
\]
which correspond to retrieving completion evidence from the image-side neighborhood with a text-side query, and the symmetric reverse process. This module outputs
\[
\mathbf{z}_i^{C}\in\mathbb{R}^{d}.
\]
The heterophily expert takes $\mathbf{H}$ and $\bar{\mathbf{A}}^{H}$ as input and, through signed polynomial filtering, produces
\[
\mathbf{z}_i^{H}\in\mathbb{R}^{d}.
\]

The outputs of the three experts are integrated by residual gating fusion. The final node representation is denoted by
\[
\mathbf{z}_i\in\mathbb{R}^{d},
\qquad
\mathbf{Z}=[\mathbf{z}_1;\dots;\mathbf{z}_N]\in\mathbb{R}^{N\times d}.
\]
The overall interface is written as
\begin{equation}
\mathbf{z}_i
=
\mathrm{Fuse}\!\left(
\mathbf{h}_i,\mathbf{z}_i^{S},\mathbf{z}_i^{C},\mathbf{z}_i^{H}
\right).
\label{eq:app_overall_fuse}
\end{equation}

Accordingly, the training objective is defined as
\begin{equation}
\mathcal{L}
=
\mathcal{L}_{\mathrm{task}}
+
\lambda_{\mathrm{qca}}\mathcal{L}_{\mathrm{qca}}
+
\lambda_{\mathrm{evi}}\mathcal{L}_{\mathrm{evi}}
+
\lambda_{\mathrm{bal}}\mathcal{L}_{\mathrm{bal}}.
\label{eq:app_overall_loss}
\end{equation}
Here $\mathcal{L}_{\mathrm{task}}$ is the main task loss, $\mathcal{L}_{\mathrm{qca}}$ is the alignment constraint for complementary completion, $\mathcal{L}_{\mathrm{evi}}$ is the evidential regularization imposed on the router, and $\mathcal{L}_{\mathrm{bal}}$ is the batch-level role balancing term. Their full definitions are given in \cref{app:fusion_objectives}.

Unless stated otherwise, the following conventions are used throughout Appendix B: the router is computed only on observed edges $\mathcal{E}$; the three routed adjacencies are used solely to characterize the propagation weights of the shared, complementary, and heterophilous channels; and the candidate neighborhood of the complementary expert is always defined over the observed neighbor set. The hyperparameter search space and the configurations used in the experiments are provided in Appendix D.

\subsection{Edge Role Router}
\label{app:edge_role_router}

This subsection corresponds to the edge role estimation with structural confidence calibration module in the main paper.

Given the modality-specific node representations
\[
\mathbf{H}^{T}=[\mathbf{h}_1^{T};\dots;\mathbf{h}_N^{T}],\qquad
\mathbf{H}^{I}=[\mathbf{h}_1^{I};\dots;\mathbf{h}_N^{I}]
\]
and the observed edge set $\mathcal{E}$, the edge role router outputs the following quantities on each observed edge $(i,j)\in\mathcal{E}$:
\[
\rho_{ij}^{T},\ \rho_{ij}^{I},\ 
\pi_{ij}^{S},\ \pi_{ij}^{C},\ \pi_{ij}^{H},\ 
\beta_{ij},\ \boldsymbol{\alpha}_{ij},\ c_{ij},\
\bar{A}_{ij}^{S},\ \bar{A}_{ij}^{C},\ \bar{A}_{ij}^{H}.
\]

\paragraph{Semantic edge feature.}
For each observed edge $(i,j)\in\mathcal{E}$, local semantic consistency is first measured in the text and image modalities. To reduce the effect of modality-specific scale differences, layer normalization is applied to node embeddings before computing cosine similarity:
\begin{equation}
s_{ij}^{T}
=
\cos\!\big(\mathrm{LN}(\mathbf{h}_i^{T}),\,\mathrm{LN}(\mathbf{h}_j^{T})\big),
\qquad
s_{ij}^{I}
=
\cos\!\big(\mathrm{LN}(\mathbf{h}_i^{I}),\,\mathrm{LN}(\mathbf{h}_j^{I})\big).
\label{eq:router_sem_score_app}
\end{equation}
This yields the three-dimensional semantic edge feature
\begin{equation}
\mathbf{v}_{ij}^{\mathrm{sem}}
=
\big[
s_{ij}^{T},\ 
s_{ij}^{I},\ 
|s_{ij}^{T}-s_{ij}^{I}|
\big]
\in\mathbb{R}^{3}.
\label{eq:router_sem_feat_app}
\end{equation}

\paragraph{Structural edge feature.}
Let the node degree be
\[
d_i=\sum_{j=1}^{N}A_{ij}.
\]
Based on this, four local structural statistics are introduced\cite{liben2007link}:
\begin{equation}
\mathrm{CN}_{ij}
=
|\mathcal{N}(i)\cap \mathcal{N}(j)|,
\label{eq:router_cn_app}
\end{equation}
\begin{equation}
\mathrm{Jacc}_{ij}
=
\frac{|\mathcal{N}(i)\cap \mathcal{N}(j)|}
{|\mathcal{N}(i)\cup \mathcal{N}(j)|},
\label{eq:router_jacc_app}
\end{equation}
\begin{equation}
\mathrm{AA}_{ij}
=
\sum_{k\in \mathcal{N}(i)\cap \mathcal{N}(j)}
\frac{1}{\log d_k},
\label{eq:router_aa_app}
\end{equation}
\begin{equation}
\mathrm{PA}_{ij}
=
d_i d_j.
\label{eq:router_pa_app}
\end{equation}
Accordingly, the structural edge feature is defined as
\begin{equation}
\mathbf{v}_{ij}^{\mathrm{str}}
=
\big[
\tilde{A}_{ij},\
\mathrm{CN}_{ij},\
\mathrm{Jacc}_{ij},\
\mathrm{AA}_{ij},\
\mathrm{PA}_{ij},\
\log(d_i+1),\
\log(d_j+1)
\big]
\in\mathbb{R}^{7}.
\label{eq:router_str_feat_app}
\end{equation}

\paragraph{Modality-wise role factors and analytic role distribution.}
Rather than applying a black-box softmax over the three roles, RoleMAG first estimates two modality-wise availability factors from the semantic edge feature:
\begin{equation}
\rho_{ij}^{T}
=
\sigma\!\big(g_{T}(\mathbf{v}_{ij}^{\mathrm{sem}})\big),
\qquad
\rho_{ij}^{I}
=
\sigma\!\big(g_{I}(\mathbf{v}_{ij}^{\mathrm{sem}})\big),
\label{eq:router_rho_app}
\end{equation}
where $g_T(\cdot)$ and $g_I(\cdot)$ are lightweight MLPs, and $\sigma(\cdot)$ denotes the sigmoid function. The three-role distribution is then given in analytic form:
\begin{equation}
\pi_{ij}^{S}
=
\rho_{ij}^{T}\rho_{ij}^{I},
\qquad
\pi_{ij}^{H}
=
(1-\rho_{ij}^{T})(1-\rho_{ij}^{I}),
\label{eq:router_pi_sh_app}
\end{equation}
\begin{equation}
\pi_{ij}^{C}
=
\rho_{ij}^{T}(1-\rho_{ij}^{I})
+
(1-\rho_{ij}^{T})\rho_{ij}^{I}.
\label{eq:router_pi_c_app}
\end{equation}
It follows directly that
\begin{equation}
\pi_{ij}^{S}+\pi_{ij}^{C}+\pi_{ij}^{H}=1.
\label{eq:router_pi_sum_app}
\end{equation}

\paragraph{Evidence strength and Dirichlet parameterization.}
After obtaining the role distribution, the router further estimates edge-level evidence strength. The semantic strength and structural support are defined as
\begin{equation}
s_{ij}^{\mathrm{sem}}
=
\mathrm{softplus}\!\big(g_{\mathrm{sem}}(\mathbf{v}_{ij}^{\mathrm{sem}})\big),
\qquad
s_{ij}^{\mathrm{str}}
=
\mathrm{softplus}\!\big(g_{\mathrm{str}}(\mathbf{v}_{ij}^{\mathrm{str}})\big),
\label{eq:router_strength_app}
\end{equation}
where $g_{\mathrm{sem}}(\cdot)$ and $g_{\mathrm{str}}(\cdot)$ are lightweight mappings. The total evidence is defined as
\begin{equation}
\beta_{ij}
=
s_{ij}^{\mathrm{sem}}\, s_{ij}^{\mathrm{str}}.
\label{eq:router_beta_app}
\end{equation}
Accordingly, the evidence vector for edge $(i,j)$ can be written as
\begin{equation}
\mathbf{e}_{ij}
=
\beta_{ij}\boldsymbol{\pi}_{ij},
\qquad
\boldsymbol{\pi}_{ij}
=
[\pi_{ij}^{S},\pi_{ij}^{C},\pi_{ij}^{H}],
\label{eq:router_evidence_vec_app}
\end{equation}
and the corresponding Dirichlet parameters are
\begin{equation}
\boldsymbol{\alpha}_{ij}
=
\mathbf{1}+\mathbf{e}_{ij}.
\label{eq:router_alpha_app}
\end{equation}
Since there are three roles in total, we have
\begin{equation}
\sum_{r\in\{S,C,H\}}\alpha_{ij}^{r}
=
3+\beta_{ij}.
\label{eq:router_alpha_sum_app}
\end{equation}

\paragraph{Confidence coefficient and effective routing weights.}
The propagation stage uses the effective confidence coefficient
\begin{equation}
c_{ij}
=
\frac{\beta_{ij}}{3+\beta_{ij}}
=
1-\frac{3}{\sum_{r\in\{S,C,H\}}\alpha_{ij}^{r}}.
\label{eq:router_confidence_app}
\end{equation}
Based on this coefficient, the effective adjacency weights of the three channels are defined as
\begin{equation}
\bar{A}_{ij}^{S}
=
c_{ij}\pi_{ij}^{S},
\qquad
\bar{A}_{ij}^{C}
=
c_{ij}\pi_{ij}^{C},
\qquad
\bar{A}_{ij}^{H}
=
c_{ij}\pi_{ij}^{H}.
\label{eq:router_effective_adj_app}
\end{equation}
For non-observed edges $(i,j)\notin\mathcal{E}$, we set
\begin{equation}
\bar{A}_{ij}^{S}=\bar{A}_{ij}^{C}=\bar{A}_{ij}^{H}=0.
\label{eq:router_nonedge_zero_app}
\end{equation}
Therefore,
\begin{equation}
\bar{A}_{ij}^{S}+\bar{A}_{ij}^{C}+\bar{A}_{ij}^{H}=c_{ij},
\qquad (i,j)\in\mathcal{E}.
\label{eq:router_effective_sum_app}
\end{equation}

\paragraph{Router interface.}
After collecting all edge-level quantities according to the indices of observed edges, the overall interface of the edge role router can be written as
\begin{equation}
\Big(
\bar{\mathbf{A}}^{S},
\bar{\mathbf{A}}^{C},
\bar{\mathbf{A}}^{H},
\mathcal{C},
\mathcal{B},
\Pi,
\mathcal{A}_{\alpha}
\Big)
=
\mathrm{Router}\!\left(
\mathbf{H}^{T},
\mathbf{H}^{I},
\tilde{\mathbf{A}},
\mathcal{E}
\right),
\label{eq:router_interface_app}
\end{equation}
where
\[
\mathcal{C}=\{c_{ij}\}_{(i,j)\in\mathcal{E}},\qquad
\mathcal{B}=\{\beta_{ij}\}_{(i,j)\in\mathcal{E}},
\]
\[
\Pi=\{\boldsymbol{\pi}_{ij}\}_{(i,j)\in\mathcal{E}},\qquad
\mathcal{A}_{\alpha}=\{\boldsymbol{\alpha}_{ij}\}_{(i,j)\in\mathcal{E}}.
\]

The structural edge feature $\mathbf{v}_{ij}^{\mathrm{str}}$ depends only on the fixed input graph and can therefore be precomputed and cached. By contrast, the semantic edge feature $\mathbf{v}_{ij}^{\mathrm{sem}}$ depends on the current-round modality representations and should be constructed dynamically during the forward pass. Under mini-batch or subgraph training, the structural statistics are still precomputed on the original graph and then retrieved by edge index.

\subsection{Shared Expert}
\label{app:shared_expert}

This subsection corresponds to the shared-channel propagation module in the main paper.

Given the shared-channel effective adjacency
\[
\bar{\mathbf{A}}^{S}\in\mathbb{R}^{N\times N}
\]
and the unified node representations
\[
\mathbf{H}=[\mathbf{h}_1;\dots;\mathbf{h}_N]\in\mathbb{R}^{N\times d},
\]
the shared expert performs role-weighted neighborhood propagation over shared and consistent neighborhoods, producing
\[
\mathbf{Z}^{S}=[\mathbf{z}_1^{S};\dots;\mathbf{z}_N^{S}]\in\mathbb{R}^{N\times d}.
\]

For any node $v_i$, the output of the shared expert is defined as
\begin{equation}
\mathbf{z}_i^{S}
=
\sum_{j\in\mathcal{N}(i)}
\bar{A}_{ij}^{S}\mathbf{W}^{S}\mathbf{h}_j,
\label{eq:shared_expert_node_app}
\end{equation}
where $\mathbf{W}^{S}\in\mathbb{R}^{d\times d}$ is the learnable linear map of the shared channel. Stacking all nodes gives the matrix form
\begin{equation}
\mathbf{Z}^{S}
=
\bar{\mathbf{A}}^{S}\mathbf{H}\mathbf{W}^{S}.
\label{eq:shared_expert_matrix_app}
\end{equation}
This propagation is restricted to edges that are routed to the shared role and supported by sufficient evidence.

From the perspective of edge-wise message passing, the message sent along edge $(j,i)$ is
\begin{equation}
\mathbf{m}_{j\rightarrow i}^{S}
=
\bar{A}_{ij}^{S}\mathbf{W}^{S}\mathbf{h}_j,
\label{eq:shared_message_app}
\end{equation}
and the node representation is obtained by summing messages from the shared neighborhood:
\begin{equation}
\mathbf{z}_i^{S}
=
\sum_{j\in\mathcal{N}(i)}
\mathbf{m}_{j\rightarrow i}^{S}.
\label{eq:shared_aggregate_app}
\end{equation}

The module interface of the shared expert can therefore be written as
\begin{equation}
\mathbf{Z}^{S}
=
\mathrm{SharedExpert}(\mathbf{H},\bar{\mathbf{A}}^{S};\mathbf{W}^{S}).
\label{eq:shared_interface_app}
\end{equation}
This interface makes explicit the weight dependency of shared-channel propagation: only edges assigned to the shared role participate in consistency-oriented propagation.

\subsection{Directional Complementary Expert}
\label{app:complementary_expert}

This subsection corresponds to the directional complementary propagation module in the main paper.

Given the complementary-channel effective adjacency
\[
\bar{\mathbf{A}}^{C}\in\mathbb{R}^{N\times N},
\]
and the modality-specific node representations
\[
\mathbf{H}^{T}=[\mathbf{h}_1^{T};\dots;\mathbf{h}_N^{T}]\in\mathbb{R}^{N\times d},
\qquad
\mathbf{H}^{I}=[\mathbf{h}_1^{I};\dots;\mathbf{h}_N^{I}]\in\mathbb{R}^{N\times d},
\]
the directional complementary expert produces
\[
\mathbf{Z}^{C}=[\mathbf{z}_1^{C};\dots;\mathbf{z}_N^{C}]\in\mathbb{R}^{N\times d}.
\]

\paragraph{Directional decomposition.}
Using the modality usability factors $\rho_{ij}^{T}$ and $\rho_{ij}^{I}$ obtained in \cref{app:edge_role_router}, two unnormalized directional strengths are first defined as
\begin{equation}
\tilde{d}_{ij}^{T\rightarrow I}
=
\rho_{ij}^{T}(1-\rho_{ij}^{I}),
\qquad
\tilde{d}_{ij}^{I\rightarrow T}
=
(1-\rho_{ij}^{T})\rho_{ij}^{I}.
\label{eq:comp_dir_unnorm_app}
\end{equation}
Let
\[
\delta_{ij}
=
\tilde{d}_{ij}^{T\rightarrow I}
+
\tilde{d}_{ij}^{I\rightarrow T}.
\]
The normalized directional weights are then defined as
\begin{equation}
d_{ij}^{T\rightarrow I}
=
\begin{cases}
\tilde{d}_{ij}^{T\rightarrow I}/\delta_{ij}, & \delta_{ij}>0,\\
0, & \text{otherwise,}
\end{cases}
\qquad
d_{ij}^{I\rightarrow T}
=
\begin{cases}
\tilde{d}_{ij}^{I\rightarrow T}/\delta_{ij}, & \delta_{ij}>0,\\
0, & \text{otherwise.}
\end{cases}
\label{eq:comp_dir_norm_app}
\end{equation}
Accordingly, for any complementary edge,
\begin{equation}
d_{ij}^{T\rightarrow I}+d_{ij}^{I\rightarrow T}=1
\qquad (\delta_{ij}>0).
\label{eq:comp_dir_sum_app}
\end{equation}
The corresponding directional effective edge weights are
\begin{equation}
\bar{A}_{ij}^{T\rightarrow I}
=
\bar{A}_{ij}^{C} d_{ij}^{T\rightarrow I},
\qquad
\bar{A}_{ij}^{I\rightarrow T}
=
\bar{A}_{ij}^{C} d_{ij}^{I\rightarrow T}.
\label{eq:comp_dir_adj_app}
\end{equation}
Therefore,
\begin{equation}
\bar{A}_{ij}^{T\rightarrow I}
+
\bar{A}_{ij}^{I\rightarrow T}
=
\bar{A}_{ij}^{C}.
\label{eq:comp_dir_adj_sum_app}
\end{equation}

\paragraph{Directional neighbor selection.}
For a center node $v_i$, the strongest $K$ complementary neighbors are retained separately in the two directions. Define
\begin{equation}
\mathcal{N}^{T\rightarrow I}(i)
=
\mathrm{TopK}_{j\in\mathcal{N}(i)}
\big(\bar{A}_{ij}^{T\rightarrow I}\big),
\qquad
\mathcal{N}^{I\rightarrow T}(i)
=
\mathrm{TopK}_{j\in\mathcal{N}(i)}
\big(\bar{A}_{ij}^{I\rightarrow T}\big),
\label{eq:comp_topk_app}
\end{equation}
where $\mathrm{TopK}$ returns the index set of the top-$K$ neighbors ranked by the corresponding directional weights in descending order. If fewer than $K$ nonzero complementary neighbors are available in one direction, all available neighbors are retained.

\paragraph{Query bottleneck.}
For any $r\in\{1,\dots,Q\}$, the direction-specific query tokens are defined as
\begin{equation}
\mathbf{q}_{i,r}^{T\rightarrow I}
=
\mathbf{q}_{r}^{T\rightarrow I}
+
\mathbf{W}_{Q}^{T}\mathbf{h}_{i}^{T},
\qquad
\mathbf{q}_{i,r}^{I\rightarrow T}
=
\mathbf{q}_{r}^{I\rightarrow T}
+
\mathbf{W}_{Q}^{I}\mathbf{h}_{i}^{I},
\label{eq:comp_query_app}
\end{equation}
where $\mathbf{q}_{r}^{T\rightarrow I},\mathbf{q}_{r}^{I\rightarrow T}\in\mathbb{R}^{d}$ are learnable queries, and $\mathbf{W}_{Q}^{T},\mathbf{W}_{Q}^{I}\in\mathbb{R}^{d\times d}$ are the query projections.

For the $T\!\rightarrow\! I$ direction, the key/value pairs are constructed from image-side neighbor representations:
\begin{equation}
\mathbf{k}_{j}^{I}
=
\mathbf{W}_{K}^{I}\mathbf{h}_{j}^{I},
\qquad
\mathbf{v}_{j}^{I}
=
\mathbf{W}_{V}^{I}\mathbf{h}_{j}^{I}.
\label{eq:comp_kv_image_app}
\end{equation}
For the $I\!\rightarrow\! T$ direction, they are symmetrically constructed from text-side neighbor representations:
\begin{equation}
\mathbf{k}_{j}^{T}
=
\mathbf{W}_{K}^{T}\mathbf{h}_{j}^{T},
\qquad
\mathbf{v}_{j}^{T}
=
\mathbf{W}_{V}^{T}\mathbf{h}_{j}^{T}.
\label{eq:comp_kv_text_app}
\end{equation}

\paragraph{Routing-aware attention bias.}
Taking $T\!\rightarrow\! I$ as an example, for any $j\in\mathcal{N}^{T\rightarrow I}(i)$, the attention logit between the $r$-th query and neighbor $v_j$ is defined as
\begin{equation}
\ell_{ij,r}^{T\rightarrow I}
=
\frac{(\mathbf{q}_{i,r}^{T\rightarrow I})^{\top}\mathbf{k}_{j}^{I}}{\sqrt{d}}
+
\lambda_{\mathrm{bias}}
\log(\bar{A}_{ij}^{T\rightarrow I}+\epsilon),
\label{eq:comp_logit_t2i_app}
\end{equation}
where $\epsilon>0$ is a numerical stability term, and $\lambda_{\mathrm{bias}}\ge 0$ is the bias scaling coefficient. A softmax normalization is then applied over the neighbor dimension:
\begin{equation}
a_{ij,r}^{T\rightarrow I}
=
\mathrm{softmax}_{j\in\mathcal{N}^{T\rightarrow I}(i)}
\big(\ell_{ij,r}^{T\rightarrow I}\big),
\label{eq:comp_attn_t2i_app}
\end{equation}
which yields the output of the $r$-th query,
\begin{equation}
\mathbf{o}_{i,r}^{T\rightarrow I}
=
\sum_{j\in\mathcal{N}^{T\rightarrow I}(i)}
a_{ij,r}^{T\rightarrow I}\mathbf{v}_{j}^{I}.
\label{eq:comp_query_out_t2i_app}
\end{equation}
In the experiments, mean pooling is used over the query dimension:
\begin{equation}
\mathbf{z}_{i}^{T\rightarrow I}
=
\frac{1}{Q}\sum_{r=1}^{Q}\mathbf{o}_{i,r}^{T\rightarrow I}.
\label{eq:comp_pool_t2i_app}
\end{equation}
The $I\!\rightarrow\! T$ direction is defined in a fully symmetric manner:
\begin{equation}
\ell_{ij,r}^{I\rightarrow T}
=
\frac{(\mathbf{q}_{i,r}^{I\rightarrow T})^{\top}\mathbf{k}_{j}^{T}}{\sqrt{d}}
+
\lambda_{\mathrm{bias}}
\log(\bar{A}_{ij}^{I\rightarrow T}+\epsilon),
\label{eq:comp_logit_i2t_app}
\end{equation}
\begin{equation}
a_{ij,r}^{I\rightarrow T}
=
\mathrm{softmax}_{j\in\mathcal{N}^{I\rightarrow T}(i)}
\big(\ell_{ij,r}^{I\rightarrow T}\big),
\qquad
\mathbf{o}_{i,r}^{I\rightarrow T}
=
\sum_{j\in\mathcal{N}^{I\rightarrow T}(i)}
a_{ij,r}^{I\rightarrow T}\mathbf{v}_{j}^{T},
\label{eq:comp_attn_out_i2t_app}
\end{equation}
\begin{equation}
\mathbf{z}_{i}^{I\rightarrow T}
=
\frac{1}{Q}\sum_{r=1}^{Q}\mathbf{o}_{i,r}^{I\rightarrow T}.
\label{eq:comp_pool_i2t_app}
\end{equation}

\paragraph{Directional completion fusion.}
The completion outputs from the two directions are first concatenated inside the expert and then projected back to the unified dimension:
\begin{equation}
\mathbf{z}_{i}^{C}
=
\mathbf{W}^{C}
\big[
\mathbf{z}_{i}^{T\rightarrow I}
\ \Vert\
\mathbf{z}_{i}^{I\rightarrow T}
\big],
\label{eq:comp_final_app}
\end{equation}
where $\mathbf{W}^{C}\in\mathbb{R}^{d\times 2d}$. The module interface of the directional complementary expert is therefore written as
\begin{equation}
\mathbf{Z}^{C}
=
\mathrm{CompExpert}
\big(
\mathbf{H}^{T},\mathbf{H}^{I},
\bar{\mathbf{A}}^{C},
\mathbf{D}^{T\rightarrow I},\mathbf{D}^{I\rightarrow T};
\Theta^{C}
\big),
\label{eq:comp_interface_app}
\end{equation}
where $\mathbf{D}^{T\rightarrow I}=[d_{ij}^{T\rightarrow I}]$, $\mathbf{D}^{I\rightarrow T}=[d_{ij}^{I\rightarrow T}]$, and $\Theta^{C}$ denotes the full set of learnable parameters in this module.

\paragraph{Boundary conditions.}
If $\mathcal{N}^{T\rightarrow I}(i)=\varnothing$, define
\[
\mathbf{z}_{i}^{T\rightarrow I}=\mathbf{0};
\]
if $\mathcal{N}^{I\rightarrow T}(i)=\varnothing$, define
\[
\mathbf{z}_{i}^{I\rightarrow T}=\mathbf{0}.
\]
If both directions are empty, then
\[
\mathbf{z}_{i}^{C}=\mathbf{0}.
\]

TopK selection is always performed separately on the complementary edge list according to directional weights, and the attention bias is injected into the logits before softmax. The search ranges and final settings of $K$, $Q$, and $\lambda_{\mathrm{bias}}$ are provided in Appendix D.

\subsection{Heterophily-aware Propagation via Signed Polynomial Filtering}
\label{app:heterophily_expert}

This subsection corresponds to the heterophily-aware propagation module in the main paper.

Given the heterophily-channel effective adjacency
\[
\bar{\mathbf{A}}^{H}\in\mathbb{R}^{N\times N},
\]
and the unified node representations
\[
\mathbf{H}=[\mathbf{h}_1;\dots;\mathbf{h}_N]\in\mathbb{R}^{N\times d},
\]
the heterophily expert produces
\[
\mathbf{Z}^{H}=[\mathbf{z}_1^{H};\dots;\mathbf{z}_N^{H}]\in\mathbb{R}^{N\times d}.
\]

\paragraph{Normalized heterophilous adjacency.}
Let the degree matrix of the heterophily channel be
\begin{equation}
\mathbf{D}^{H}_{ii}
=
\sum_{j=1}^{N}\bar{A}_{ij}^{H}.
\label{eq:hetero_degree_app}
\end{equation}
The corresponding symmetrically normalized heterophilous adjacency is defined as
\begin{equation}
\tilde{\mathbf{A}}^{H}
=
(\mathbf{D}^{H})^{-\frac12}\bar{\mathbf{A}}^{H}(\mathbf{D}^{H})^{-\frac12}.
\label{eq:hetero_norm_adj_app}
\end{equation}
If $\mathbf{D}^{H}_{ii}=0$, the corresponding inverse square-root entry is set to $0$ for numerical stability.

\paragraph{Signed polynomial filter.}
Define
\[
\mathbf{S}_{0}=\mathbf{H},\qquad
\mathbf{S}_{r}=\tilde{\mathbf{A}}^{H}\mathbf{S}_{r-1}\quad (r\ge 1).
\]
The general form of the signed polynomial filter is then
\begin{equation}
\mathbf{Z}^{H}
=
\sum_{r=0}^{R}\gamma_r\mathbf{S}_{r},
\label{eq:hetero_poly_general_app}
\end{equation}
where $\gamma_r\in\mathbb{R}$ is a learnable scalar. The current implementation adopts a low-order setting with $R=2$, namely
\begin{equation}
\mathbf{Z}^{H}
=
\gamma_0\mathbf{H}
+
\gamma_1\tilde{\mathbf{A}}^{H}\mathbf{H}
+
\gamma_2(\tilde{\mathbf{A}}^{H})^2\mathbf{H}.
\label{eq:hetero_poly_matrix_app}
\end{equation}
For any node $v_i$, the corresponding output is
\begin{equation}
\mathbf{z}_i^{H}
=
\gamma_0\mathbf{h}_i
+
\gamma_1\sum_{j=1}^{N}\tilde{A}_{ij}^{H}\mathbf{h}_j
+
\gamma_2\sum_{k=1}^{N}\big[(\tilde{\mathbf{A}}^{H})^2\big]_{ik}\mathbf{h}_k.
\label{eq:hetero_poly_node_app}
\end{equation}

Here $\gamma_0$ controls the $0$-hop retention term within the heterophily channel, $\gamma_1$ governs the first-order heterophilous signal, and $\gamma_2$ captures the second-order relay effect. These coefficients are not constrained to be nonnegative, nor are they required to sum to $1$.

\paragraph{Module interface and boundary cases.}
The module interface of the heterophily expert is written as
\begin{equation}
\mathbf{Z}^{H}
=
\mathrm{HeteroExpert}
\big(
\mathbf{H},\bar{\mathbf{A}}^{H};\gamma_0,\gamma_1,\gamma_2
\big).
\label{eq:hetero_interface_app}
\end{equation}
If node $v_i$ has no available neighbors in the heterophily channel, the $i$-th row of \cref{eq:hetero_norm_adj_app} becomes all zeros, and thus
\begin{equation}
\mathbf{z}_i^{H}
=
\gamma_0\mathbf{h}_i.
\label{eq:hetero_isolated_app}
\end{equation}

In computation, \cref{eq:hetero_poly_matrix_app} can be implemented recursively. One first computes
\[
\mathbf{S}_1=\tilde{\mathbf{A}}^{H}\mathbf{H},
\]
and then
\[
\mathbf{S}_2=\tilde{\mathbf{A}}^{H}\mathbf{S}_1,
\]
thereby avoiding the explicit construction of the dense matrix $(\tilde{\mathbf{A}}^{H})^2$.

\subsection{Residual Gating Fusion and Training Objectives}
\label{app:fusion_objectives}

This subsection corresponds to the residual gating fusion and the overall training objective in the main paper.

Given the outputs of the three propagation channels
\[
\mathbf{Z}^{S}=[\mathbf{z}_1^{S};\dots;\mathbf{z}_N^{S}],\quad
\mathbf{Z}^{C}=[\mathbf{z}_1^{C};\dots;\mathbf{z}_N^{C}],\quad
\mathbf{Z}^{H}=[\mathbf{z}_1^{H};\dots;\mathbf{z}_N^{H}],
\]
and the unified input representation
\[
\mathbf{H}=[\mathbf{h}_1;\dots;\mathbf{h}_N]\in\mathbb{R}^{N\times d},
\]
RoleMAG adaptively fuses the three expert channels through a lightweight gating network, while preserving an explicit residual path from the center representation. The final output is denoted by
\[
\mathbf{Z}=[\mathbf{z}_1;\dots;\mathbf{z}_N]\in\mathbb{R}^{N\times d}.
\]

\paragraph{Residual gating fusion.}
For any node $v_i$, the fusion input is first constructed as
\begin{equation}
\mathbf{u}_i
=
[\mathbf{h}_i \,\Vert\, \mathbf{z}_i^{S} \,\Vert\, \mathbf{z}_i^{C} \,\Vert\, \mathbf{z}_i^{H}]
\in\mathbb{R}^{4d},
\label{eq:fusion_input_app}
\end{equation}
The gating MLP then produces a three-dimensional logit vector:
\begin{equation}
\boldsymbol{\ell}_i^{g}
=
\mathrm{MLP}_{g}(\mathbf{u}_i)
\in\mathbb{R}^{3}.
\label{eq:gating_logits_app}
\end{equation}
The corresponding fusion weights are defined as
\begin{equation}
[g_i^{S},g_i^{C},g_i^{H}]
=
\operatorname{softmax}(\boldsymbol{\ell}_i^{g}),
\label{eq:gating_weights_app}
\end{equation}
The final representation is given by
\begin{equation}
\mathbf{z}_i
=
\mathbf{h}_i
+
g_i^{S}\mathbf{z}_i^{S}
+
g_i^{C}\mathbf{z}_i^{C}
+
g_i^{H}\mathbf{z}_i^{H}.
\label{eq:residual_gating_fusion_app}
\end{equation}
where $\mathbf{h}_i$ serves as a global residual anchor and does not participate in the softmax competition among the gates.

\paragraph{Fusion module interface.}
Accordingly, the fusion module can be written as
\begin{equation}
\mathbf{Z}
=
\mathrm{Fuse}
\big(
\mathbf{H},\mathbf{Z}^{S},\mathbf{Z}^{C},\mathbf{Z}^{H}
\big).
\label{eq:fusion_interface_app}
\end{equation}
The output $\mathbf{Z}$ is then used as the input to the downstream task head.

\paragraph{Task head and main task supervision.}
In the main experiments, RoleMAG is instantiated on two graph-centric tasks, namely node classification and link prediction. For node classification, the task head is defined as
\begin{equation}
\hat{\mathbf{y}}_i
=
\psi_{\mathrm{task}}(\mathbf{z}_i),
\label{eq:task_head_node_app}
\end{equation}
and the corresponding main-task loss is written as
\begin{equation}
\mathcal{L}_{\mathrm{task}}^{\mathrm{node}}
=
\frac{1}{|\mathcal{V}_L|}
\sum_{i\in\mathcal{V}_{L}}
\mathrm{CE}(\hat{\mathbf{y}}_i,y_i).
\label{eq:task_loss_node_app}
\end{equation}

For link prediction, the task head operates on node pairs:
\begin{equation}
\hat{s}_{ij}
=
\psi_{\mathrm{task}}(\mathbf{z}_i,\mathbf{z}_j).
\label{eq:task_head_pair_app}
\end{equation}
The corresponding
\[
\mathcal{L}_{\mathrm{task}}^{\mathrm{link}}
\]
is instantiated by the binary classification loss or ranking loss used in the experiment. At the method level, it is uniformly denoted by $\mathcal{L}_{\mathrm{task}}$. The task-specific settings are provided in Appendix D.

\paragraph{Cross-modal completion alignment.}
To ensure that the complementary expert learns directionally correct cross-modal completion, a contrastive alignment constraint is imposed on the completion outputs of both directions. Taking the $T\rightarrow I$ direction as an example, define
\begin{equation}
\mathbf{u}_i^{T\rightarrow I}
=
p_{I}(\mathbf{z}_i^{T\rightarrow I}),
\qquad
\mathbf{v}_i^{I}
=
p_{I}(\mathbf{h}_i^{I}),
\label{eq:qca_projection_t2i_app}
\end{equation}
where $p_I(\cdot)$ is the projection head that maps representations into the image-side alignment space. The corresponding InfoNCE objective is
\begin{equation}
\mathcal{L}_{\mathrm{qca}}^{T\rightarrow I}
=
-
\frac{1}{|\Omega|}
\sum_{i\in\Omega}
\log
\frac{
\exp\!\big(
\cos(\mathbf{u}_i^{T\rightarrow I},\mathbf{v}_i^{I})/\tau
\big)
}{
\sum_{j\in\Omega}
\exp\!\big(
\cos(\mathbf{u}_i^{T\rightarrow I},\mathbf{v}_j^{I})/\tau
\big)
},
\label{eq:qca_loss_t2i_app}
\end{equation}
where $\Omega$ denotes the set of nodes participating in the contrastive objective within the current mini-batch, and $\tau$ is the temperature parameter. Symmetrically,
\begin{equation}
\mathbf{u}_i^{I\rightarrow T}
=
p_{T}(\mathbf{z}_i^{I\rightarrow T}),
\qquad
\mathbf{v}_i^{T}
=
p_{T}(\mathbf{h}_i^{T}),
\label{eq:qca_projection_i2t_app}
\end{equation}
and define
\begin{equation}
\mathcal{L}_{\mathrm{qca}}^{I\rightarrow T}
=
-
\frac{1}{|\Omega|}
\sum_{i\in\Omega}
\log
\frac{
\exp\!\big(
\cos(\mathbf{u}_i^{I\rightarrow T},\mathbf{v}_i^{T})/\tau
\big)
}{
\sum_{j\in\Omega}
\exp\!\big(
\cos(\mathbf{u}_i^{I\rightarrow T},\mathbf{v}_j^{T})/\tau
\big)
}.
\label{eq:qca_loss_i2t_app}
\end{equation}
Finally,
\begin{equation}
\mathcal{L}_{\mathrm{qca}}
=
\mathcal{L}_{\mathrm{qca}}^{T\rightarrow I}
+
\mathcal{L}_{\mathrm{qca}}^{I\rightarrow T}.
\label{eq:qca_loss_total_app}
\end{equation}

\paragraph{Evidential regularization.}
In the router, the uncertainty of edge $(i,j)$ is already characterized by $\beta_{ij}$, $\boldsymbol{\alpha}_{ij}$, and $c_{ij}$. During training, two edge sets are further distinguished: $\mathcal{E}_{\mathrm{in}}$ denotes the observed real edges, whereas $\mathcal{E}_{\mathrm{out}}$ denotes pseudo edges constructed by random rewiring. We first define the confidence-separation term
\begin{equation}
\mathcal{L}_{\mathrm{evi}}^{\mathrm{conf}}
=
-
\frac{1}{|\mathcal{E}_{\mathrm{in}}|}
\sum_{(i,j)\in\mathcal{E}_{\mathrm{in}}}
\log c_{ij}
-
\frac{1}{|\mathcal{E}_{\mathrm{out}}|}
\sum_{(i,j)\in\mathcal{E}_{\mathrm{out}}}
\log(1-c_{ij}).
\label{eq:evi_conf_app}
\end{equation}
In addition, a Dirichlet-space regularization term is imposed on the pseudo edges:
\begin{equation}
\mathcal{L}_{\mathrm{evi}}^{\mathrm{KL}}
=
\frac{1}{|\mathcal{E}_{\mathrm{out}}|}
\sum_{(i,j)\in\mathcal{E}_{\mathrm{out}}}
\mathrm{KL}
\Big[
\mathrm{Dir}(\boldsymbol{\alpha}_{ij})
\;\Vert\;
\mathrm{Dir}(\mathbf{1})
\Big],
\label{eq:evi_kl_app}
\end{equation}
where $\mathbf{1}\in\mathbb{R}^{3}$ denotes the uniform prior over the three roles\cite{malinin2018prior,sensoy2018evidential}. Therefore,
\begin{equation}
\mathcal{L}_{\mathrm{evi}}
=
\mathcal{L}_{\mathrm{evi}}^{\mathrm{conf}}
+
\eta_{\mathrm{KL}}\mathcal{L}_{\mathrm{evi}}^{\mathrm{KL}}.
\label{eq:evi_total_app}
\end{equation}

\paragraph{Role balancing regularization.}
To avoid collapse to a single role in the early stage of training, the soft assignment mass of each role is defined over the edge set $\mathcal{E}_{b}$ of the current mini-batch:
\begin{equation}
Imp_r
=
\sum_{(i,j)\in\mathcal{E}_{b}}
\pi_{ij}^{r},
\qquad
r\in\{S,C,H\}.
\label{eq:role_importance_app}
\end{equation}
Let
\begin{equation}
\bar{Imp}
=
\frac{1}{3}(Imp_S+Imp_C+Imp_H),
\label{eq:mean_role_importance_app}
\end{equation}
Then the coefficient of variation is written as
\begin{equation}
\mathrm{CV}\!\left(\{Imp_S,Imp_C,Imp_H\}\right)
=
\frac{
\sqrt{
\frac{1}{3}\sum_{r\in\{S,C,H\}}(Imp_r-\bar{Imp})^2
}
}{
\bar{Imp}+\epsilon
}.
\label{eq:cv_role_mass_app}
\end{equation}
Accordingly, the role balancing loss is defined as
\begin{equation}
\mathcal{L}_{\mathrm{bal}}
=
\mathrm{CV}\!\left(\{Imp_S,Imp_C,Imp_H\}\right)^2.
\label{eq:role_balance_app}
\end{equation}

\paragraph{Overall objective.}
Combining the main-task supervision, completion alignment, evidential regularization, and role balancing, the final training objective is
\begin{equation}
\mathcal{L}
=
\mathcal{L}_{\mathrm{task}}
+
\lambda_{\mathrm{evi}}\mathcal{L}_{\mathrm{evi}}
+
\lambda_{\mathrm{qca}}\mathcal{L}_{\mathrm{qca}}
+
\lambda_{\mathrm{bal}}\mathcal{L}_{\mathrm{bal}},
\label{eq:overall_loss_app}
\end{equation}
where $\lambda_{\mathrm{evi}},\lambda_{\mathrm{qca}},\lambda_{\mathrm{bal}}\ge 0$ are hyperparameters. The search ranges and final values of the temperature $\tau$, the KL balance coefficient $\eta_{\mathrm{KL}}$, and all auxiliary loss coefficients are provided in Appendix D.4.

During training, the full forward pipeline is executed and all four loss terms in \cref{eq:overall_loss_app} are computed jointly. During inference, only forward propagation is retained, and $\mathcal{L}_{\mathrm{qca}}$, $\mathcal{L}_{\mathrm{evi}}$, and $\mathcal{L}_{\mathrm{bal}}$ are no longer evaluated.

\subsection{Complexity and Design Discussion}
\label{app:complexity_discussion}

This subsection corresponds to the efficiency analysis and complexity decomposition in the main paper.

Let the number of nodes be $N=|\mathcal{V}|$, the number of observed edges be $M=|\mathcal{E}|$, the hidden dimension be $d$, the number of retained neighbors per direction be $K$, the number of queries be $Q$, and the filter order of the heterophily expert be $R$.

\paragraph{Edge role router.}
Assuming that the structural edge feature $\mathbf{v}_{ij}^{\mathrm{str}}$ has been precomputed and cached, the online cost of the router mainly comes from semantic edge-feature construction and the lightweight MLP. For each observed edge, computing
\[
s_{ij}^{T},\ s_{ij}^{I},\ |s_{ij}^{T}-s_{ij}^{I}|
\]
has complexity
\begin{equation}
\mathcal{O}(Md),
\label{eq:complex_router_sem_app}
\end{equation}
while the lightweight mappings and evidential parameterization can be written as
\begin{equation}
\mathcal{O}(M).
\label{eq:complex_router_mlp_app}
\end{equation}
Therefore, the dominant forward complexity of the router per iteration is
\begin{equation}
\mathcal{O}(Md).
\label{eq:complex_router_total_app}
\end{equation}

\paragraph{Shared expert.}
The core computation of the shared expert is
\[
\mathbf{Z}^{S}=\bar{\mathbf{A}}^{S}\mathbf{H}\mathbf{W}^{S}.
\]
If the linear projection and sparse aggregation are counted separately, the complexity becomes
\begin{equation}
\mathcal{O}(Nd^{2})+\mathcal{O}(Md).
\label{eq:complex_shared_app}
\end{equation}

\paragraph{Directional complementary expert.}
The cost of this module mainly comes from three parts. The cost of directional decomposition and directional edge-weight construction is
\begin{equation}
\mathcal{O}(M).
\label{eq:complex_comp_dir_app}
\end{equation}
The node-wise TopK selection can be written as
\begin{equation}
\mathcal{O}(M\log K),
\label{eq:complex_comp_topk_app}
\end{equation}
With a linear-time selection implementation, this term can be further reduced toward $\mathcal{O}(M)$. The dominant term of the query-bottleneck retrieval over the two directions is
\begin{equation}
\mathcal{O}(NQKd),
\label{eq:complex_comp_attn_app}
\end{equation}
together with the cost of the query, key, and value projections:
\begin{equation}
\mathcal{O}(Nd^{2}).
\label{eq:complex_comp_proj_app}
\end{equation}
Hence, the total complexity of the complementary expert is
\begin{equation}
\mathcal{O}(M\log K)+\mathcal{O}(NQKd)+\mathcal{O}(Nd^{2}).
\label{eq:complex_comp_total_app}
\end{equation}

\paragraph{Heterophily expert.}
From \cref{eq:hetero_poly_general_app},
\[
\mathbf{Z}^{H}
=
\sum_{r=0}^{R}\gamma_r(\tilde{\mathbf{A}}^{H})^r\mathbf{H}.
\]
if this term is computed recursively, its complexity is
\begin{equation}
\mathcal{O}(RMd).
\label{eq:complex_hetero_app}
\end{equation}
In the current implementation, $R=2$. The propagation cost of the heterophily expert is therefore
\begin{equation}
\mathcal{O}(2Md).
\label{eq:complex_hetero_r2_app}
\end{equation}

\paragraph{Fusion and auxiliary objectives.}
The forward pass of residual gating fusion mainly consists of a gating MLP with input dimension $4d$, and its cost is therefore
\begin{equation}
\mathcal{O}(Nd^{2}).
\label{eq:complex_fusion_app}
\end{equation}
During training, the auxiliary losses introduce two additional types of cost. $\mathcal{L}_{\mathrm{qca}}$ forms contrastive similarities over the mini-batch node set $\Omega$, whose standard implementation has complexity
\begin{equation}
\mathcal{O}(|\Omega|^{2}d),
\label{eq:complex_qca_app}
\end{equation}
Meanwhile, $\mathcal{L}_{\mathrm{evi}}$ performs certainty matching and Dirichlet regularization over the pseudo-edge set $\mathcal{E}_{\mathrm{out}}$, which introduces additional complexity
\begin{equation}
\mathcal{O}(|\mathcal{E}_{\mathrm{out}}|).
\label{eq:complex_evi_app}
\end{equation}
By comparison, $\mathcal{L}_{\mathrm{bal}}$ only accumulates the soft role mass over the mini-batch edge set, with cost
\begin{equation}
\mathcal{O}(|\mathcal{E}_{b}|).
\label{eq:complex_bal_app}
\end{equation}

\paragraph{Overall complexity.}
Combining \cref{eq:complex_router_total_app,eq:complex_shared_app,eq:complex_comp_total_app,eq:complex_hetero_app,eq:complex_fusion_app,eq:complex_qca_app,eq:complex_evi_app,eq:complex_bal_app}, the dominant training complexity of RoleMAG per iteration can be written as
\begin{equation}
\mathcal{O}\big((R+2)Md + M\log K + NQKd + Nd^{2} + |\Omega|^{2}d + |\mathcal{E}_{\mathrm{out}}|\big).
\label{eq:complex_train_simplified_app}
\end{equation}
If the auxiliary losses used only during training are omitted, the inference complexity reduces to
\begin{equation}
\mathcal{O}\big((R+2)Md + M\log K + NQKd + Nd^{2}\big).
\label{eq:complex_infer_app}
\end{equation}

\paragraph{Memory complexity.}
Beyond the modality encoders, the additional memory of RoleMAG mainly comes from three parts. The first is the three routed edge-weight matrices
\[
\bar{\mathbf{A}}^{S},\ \bar{\mathbf{A}}^{C},\ \bar{\mathbf{A}}^{H},
\]
When stored as sparse edge lists, their memory cost is
\begin{equation}
\mathcal{O}(M).
\label{eq:memory_router_app}
\end{equation}
The second part is the TopK neighbor indices, directional weights, and attention coefficients retained in the complementary expert, whose scale is
\begin{equation}
\mathcal{O}(NQK).
\label{eq:memory_comp_app}
\end{equation}
The third part consists of the intermediate node representations of the three experts and the recursive cache of the heterophily filter, whose scale is
\begin{equation}
\mathcal{O}(Nd).
\label{eq:memory_hidden_app}
\end{equation}
Therefore, excluding encoder activations, the additional memory of RoleMAG can be summarized as
\begin{equation}
\mathcal{O}(M + NQK + Nd).
\label{eq:memory_total_app}
\end{equation}

\paragraph{Design scope.}
The current implementation follows three basic constraints. First, the router estimates roles only on the observed edges $\mathcal{E}$ and does not explicitly reconstruct a new dense topology. RoleMAG therefore focuses on how the existing neighborhood should participate in propagation, rather than on rediscovering missing relations. Second, the complementary expert restricts candidate complementary neighbors to direction-specific TopK subsets and performs cross-modal retrieval through a query bottleneck. This design controls both computation and memory, while constraining completion to local neighborhoods with relatively high confidence. Third, the heterophily expert adopts a low-order signed polynomial filter instead of a higher-order or adaptive-order spectral filter, and thus mainly captures heterophilous relay effects from local to medium ranges. The present formulation directly targets the text--image bimodal setting. If extended to three or more modalities, both the complementary direction set and the role algebra would need to be redefined.
}

\newpage
\section{Pseudocode}
\label{app:pseudocode}


\SetKwInOut{Input}{Input}
\SetKwInOut{Output}{Output}

\begin{algorithm}[htbp]
\caption{Training Procedure of RoleMAG}
\label{alg:rolemag_train}
\DontPrintSemicolon
\Input{
multimodal graph $G=(V,E,\{\mathbf X^{(m)}\}_{m\in\{T,I\}})$;
maximum epochs $E_{\max}$
}
\Output{trained parameters $\Theta$}

Initialize model parameters $\Theta$\;
Compute the normalized adjacency $\tilde{\mathbf A}$\;
Precompute and cache structural edge statistics on the observed edge set $\mathcal E$\;

\For{$e \leftarrow 1$ \KwTo $E_{\max}$}{
    Encode the two modalities to obtain $\mathbf H^{T}$ and $\mathbf H^{I}$\;
    Compute the unified input representation $\mathbf H$ via Eq.~(1)\;

    $(\mathbf Z,\mathbf Z^{T\rightarrow I},\mathbf Z^{I\rightarrow T},
    \mathcal C,\mathcal B,\Pi,\mathcal A_{\alpha})
    \leftarrow
    \textsc{RoleAwarePropagation}(\mathbf H,\mathbf H^{T},\mathbf H^{I},\tilde{\mathbf A},\mathcal E)$\;

    Compute task prediction $\hat{\mathbf Y}=\psi_{\mathrm{task}}(\mathbf Z)$ and the main loss $\mathcal L_{\mathrm{task}}$\;
    Compute $\mathcal L_{\mathrm{qca}}$ from $\mathbf Z^{T\rightarrow I},\mathbf Z^{I\rightarrow T},\mathbf H^{T},\mathbf H^{I}$ according to Appendix~B.6\;
    Compute $\mathcal L_{\mathrm{evi}}$ from $\mathcal C,\mathcal B,\mathcal A_{\alpha}$ according to Appendix~B.6\;
    Compute $\mathcal L_{\mathrm{bal}}$ from $\Pi$ according to Appendix~B.6\;

    $\mathcal L \leftarrow
    \mathcal L_{\mathrm{task}}
    + \lambda_{\mathrm{qca}}\mathcal L_{\mathrm{qca}}
    + \lambda_{\mathrm{evi}}\mathcal L_{\mathrm{evi}}
    + \lambda_{\mathrm{bal}}\mathcal L_{\mathrm{bal}}$\;

    Update $\Theta$ by back-propagation on $\mathcal L$\;
}
\Return{$\Theta$}\;
\end{algorithm}


\begin{algorithm}[htbp]
\caption{Role-aware Propagation of RoleMAG}
\label{alg:rolemag_propagation}
\DontPrintSemicolon
\Input{
unified representation $\mathbf H$;
modality-specific representations $\mathbf H^{T},\mathbf H^{I}$;
normalized adjacency $\tilde{\mathbf A}$;
observed edge set $E$
}
\Output{
final representation $\mathbf Z$;
directional complementary outputs $\mathbf Z^{T\rightarrow I},\mathbf Z^{I\rightarrow T}$;
router statistics $\mathcal C,\mathcal B,\Pi,\mathcal A_{\alpha}$
}

\tcc{1. Edge role router}
$(\bar{\mathbf A}^{S},\bar{\mathbf A}^{C},\bar{\mathbf A}^{H},
\mathcal C,\mathcal B,\Pi,\mathcal A_{\alpha})
\leftarrow
\textsc{Router}(\mathbf H^{T},\mathbf H^{I},\tilde{\mathbf A},\mathcal E)$\;

\tcc{2. Three role-aware experts}
$\mathbf Z^{S} \leftarrow \textsc{SharedExpert}(\mathbf H,\bar{\mathbf A}^{S};\mathbf W^{S})$\;

\tcc{The complementary expert internally performs directional decomposition,}
\tcc{TopK neighbor selection, query-based completion, and directional fusion}
$(\mathbf Z^{C},\mathbf Z^{T\rightarrow I},\mathbf Z^{I\rightarrow T})
\leftarrow
\textsc{CompExpert}(\mathbf H^{T},\mathbf H^{I},\bar{\mathbf A}^{C};\Theta^{C})$\;

$\mathbf Z^{H} \leftarrow \textsc{HeteroExpert}(\mathbf H,\bar{\mathbf A}^{H};\gamma_{0},\gamma_{1},\gamma_{2})$\;

\tcc{3. Residual gating fusion}
\ForEach{$v_i \in V$}{
    $\mathbf u_i \leftarrow [\mathbf h_i \,\|\, \mathbf z_i^{S} \,\|\, \mathbf z_i^{C} \,\|\, \mathbf z_i^{H}]$\;
    $[g_i^{S},g_i^{C},g_i^{H}] \leftarrow \mathrm{softmax}(\mathrm{MLP}_{g}(\mathbf u_i))$\;
    $\mathbf z_i \leftarrow
    \mathbf h_i + g_i^{S}\mathbf z_i^{S} + g_i^{C}\mathbf z_i^{C} + g_i^{H}\mathbf z_i^{H}$\;
}
$\mathbf Z \leftarrow [\mathbf z_1;\ldots;\mathbf z_N]$\;

\Return{$\mathbf Z,\mathbf Z^{T\rightarrow I},\mathbf Z^{I\rightarrow T},
\mathcal C,\mathcal B,\Pi,\mathcal A_{\alpha}$}\;
\end{algorithm}
\section{Detailed Experimental Setup}
\label{app:setup}

\subsection{Datasets and Data Splits}
\label{app:datasets_splits}

The main-paper results involve three graph-centric datasets: Toys, RedditS, and Bili\_Dance. All three are drawn from OpenMAG~\cite{wan2026openmag}. Toys and RedditS are used for node classification, while Bili\_Dance is used for link prediction. Their statistics are summarized in \cref{tab:appendix_dataset_stats}.

\begin{table*}[t]
\centering
\small
\setlength{\tabcolsep}{6pt}
\renewcommand{\arraystretch}{1.12}
\begin{tabular}{lcccccc}
\toprule
\textbf{Dataset} & \textbf{Task} & \textbf{\#Nodes} & \textbf{\#Edges} & \textbf{\#Classes} & \textbf{Image Dim.} & \textbf{Text Dim.} \\
\midrule
Toys & Node Classification & 20{,}695 & 63{,}443 & 18 & 768 & 768 \\
RedditS & Node Classification & 15{,}894 & 283{,}080 & 20 & 768 & 768 \\
Bili\_Dance & Link Prediction & 2{,}307 & 9{,}127 & -- & 768 & 768 \\
\bottomrule
\end{tabular}
\caption{\textbf{Datasets used in the main-paper graph-centric experiments.} Bili\_Dance is used for link prediction and therefore does not involve node classes.}
\label{tab:appendix_dataset_stats}
\end{table*}

All three datasets use frozen bimodal node representations as input, without end-to-end updates to the underlying multimodal encoders. Concretely, both text and image representations are extracted through an offline pre-encoding pipeline and aligned to the same representation dimension before entering the graph model. All graph-centric experiments in the main paper are built on this setting: bimodal node features are fixed first, and task-related propagation and fusion parameters are then learned over the graph structure.

For node classification, the training, validation, and test sets are randomly split from the node set according to
\[
60\% / 20\% / 20\%
\]
respectively. For link prediction, the training, validation, and test edges are randomly split from the observed edge set according to
\[
70\% / 10\% / 20\%
\]
respectively. During validation and testing, each positive edge is paired with $1000$ negative edges to compute MRR and Hits@$K$. Unless otherwise noted, all splits are generated under fixed random seeds and follow the graph-centric evaluation protocol of OpenMAG~\cite{wan2026openmag}.

\subsection{Feature Construction and Graph Preprocessing}
\label{app:preprocess}

Let the frozen representations of the visual and textual modalities be
\[
\mathbf{X}^{I}\in\mathbb{R}^{N\times d_I},
\qquad
\mathbf{X}^{T}\in\mathbb{R}^{N\times d_T}.
\]
In the current experiments, both are $768$-dimensional and are concatenated column-wise to form the node input matrix
\[
\mathbf{X}=[\mathbf{X}^{T}\Vert\mathbf{X}^{I}]
\in\mathbb{R}^{N\times(d_T+d_I)}.
\]
If the cached raw features contain missing or non-finite values, numerical cleaning is performed before they are fed into the graph model. This step is used only to ensure numerical stability and does not alter the graph structure or modal semantics.

For node classification, the graph is treated as undirected and augmented with self-loops before mini-batch sampling. Accordingly, propagation is performed over a normalized adjacency that includes reverse edges and self-loops. For link prediction, the sparse adjacency is constructed only from the training-edge subset, without adding self-loops, while preserving the edge directions induced by the split. Both tasks follow the graph-centric data-processing protocol of OpenMAG, but the routing and three-expert propagation in RoleMAG operate only on the observed edges defined by the current task~\cite{wan2026openmag}. Mini-batches for node classification are constructed through neighborhood sampling, whereas mini-batches for link prediction are organized around training edges and their local neighborhoods.

\subsection{Unified Benchmark Protocol and Baselines}
\label{app:benchmark_protocol}

The main-paper results are all obtained under the graph-centric protocol of OpenMAG. For node classification, the evaluation metrics are accuracy and macro-F1. For link prediction, the evaluation metrics are MRR and Hits@$K$. The main table in the paper reports MRR and Hits@3, while the supplementary material retains Hits@1/3/10. Unless otherwise stated, all results are reported as percentages, with mean and standard deviation computed over multiple independent runs.

The compared methods cover three representative groups of models. The first group consists of general GNN backbones, such as GCN and GAT. The second group includes classical multimodal graph learning methods, such as MMGCN, MGAT, and LGMRec. The third group contains structure-enhanced or query-driven models that are closer to the present problem setting, including DGF, DMGC, NTSFormer, Graph4MM, and TMTE~\cite{wei2019mmgcn,tao2020mgat,guo2024lgmrec,ning2025graph4mm,zhu2026tmte}. All methods are compared under the same task splits, input modalities, and evaluation criteria.

\subsection{RoleMAG Instantiation in Main Experiments}
\label{app:rolemag_impl_setup}

RoleMAG in the main paper is instantiated around an edge-role router, three role-aware propagation channels, and residual gating fusion; the full specification is given in \cref{app:method_spec,app:pseudocode}. In the main experiments, the bimodal node representations are first mapped into a unified hidden space, after which role-aware propagation is performed over the observed graph defined by the task. The router outputs the assignment strengths of three roles, namely shared, complementary, and heterophilous, on observed edges. The three experts then model consistent support, directional cross-modal completion, and heterophilous interaction, respectively, producing the node representations used for downstream prediction.

Under the main experimental setting, the shared channel captures locally consistent support across modalities; the complementary channel retains directional decomposition, TopK neighbor selection, a query bottleneck, and routing-aware attention bias; and the heterophily channel adopts low-order signed polynomial filtering to model the high-frequency interactions introduced by heterophilous edges. Concretely, the hidden dimension is set to $d=256$; the complementary channel keeps $K=16$ candidate neighbors for each direction and uses $Q=4$ query tokens; and the heterophily channel uses a second-order signed polynomial form. The key hyperparameters are listed in \cref{tab:appendix_rolemag_hparams}.

\begin{table*}[t]
\centering
\small
\setlength{\tabcolsep}{6pt}
\renewcommand{\arraystretch}{1.12}
\begin{tabular}{p{0.26\textwidth} p{0.18\textwidth} p{0.42\textwidth}}
\toprule
\textbf{Component} & \textbf{Symbol / Hyperparameter} & \textbf{Value Used in Current Experiments} \\
\midrule
Node representation dimension & $d$ & $256$ \\
Dropout & $p$ & $0.2$ \\
Router hidden dimension & $d_{\mathrm{router}}$ & $64$ \\
Complementary TopK size & $K$ & $16$ \\
Number of query tokens & $Q$ & $4$ \\
Attention bias scale & $\lambda_{\mathrm{bias}}$ & $1.0$ \\
Role balancing coefficient & $\lambda_{\mathrm{bal}}$ & $0.1$ \\
Cross-modal completion coefficient & $\lambda_{\mathrm{qca}}$ & $0.5$ \\
Contrastive temperature & $\tau$ & $0.07$ \\
Signed polynomial initialization & $(\gamma_0,\gamma_1,\gamma_2)$ & $(1.0,-0.5,0.5)$ \\
Shared propagation depth & -- & $3$ \\
Pseudo-edge ratio for $\mathcal{L}_{\mathrm{evi}}$ & out-edge ratio & $0.05$ \\
Evidential regularization coefficient & $\lambda_{\mathrm{evi}}$ & $0.1$ \\
\bottomrule
\end{tabular}
\caption{\textbf{Key hyperparameters of RoleMAG used in the main experiments.}}
\label{tab:appendix_rolemag_hparams}
\end{table*}

During training and diagnosis, we mainly record task metrics and routing statistics on observed edges. Consistent with the notation in the main paper and Appendix B, the relevant hyperparameters are denoted by $\lambda_{\mathrm{evi}}$, $\lambda_{\mathrm{bal}}$, $\lambda_{\mathrm{qca}}$, $\tau$, and $\eta_{\mathrm{KL}}$.

\subsection{Optimization and Mini-batch Training}
\label{app:optimization}

For node classification, Adam is used as the optimizer, with a maximum of $30$ training epochs, batch size $1024$, weight decay $10^{-5}$, and learning rate $5\times10^{-3}$. During training, mini-batch subgraphs are obtained through neighborhood sampling, with at most $25$ sampled neighbors per hop. The sampling depth is aligned with the number of propagation layers in the backbone graph model. Validation and test metrics are recorded under the fixed training schedule.

For link prediction, Adam is likewise used as the optimizer, with a maximum of $50$ training epochs, batch size $512$, learning rate $10^{-3}$, and weight decay $10^{-5}$. The link scorer is implemented as a three-layer MLP with hidden dimension $256$ and dropout $0.02$. During training, subgraphs are built through local neighborhood sampling, and negative edges are matched online for each positive edge to form the binary contrastive objective.

For RoleMAG, the training objective consists of the task loss, the cross-modal completion constraint, evidential regularization, and the role-balancing term; its full form is given in \cref{app:fusion_objectives}. The experiment logs mainly record task metrics together with routing statistics on observed edges, including $\beta_{ij}$, $c_{ij}$, $\boldsymbol{\pi}_{ij}$, and the evolution of node-level gates.

\subsection{Environment and Reproducibility}
\label{app:environment_commands}

All experiments are conducted under frozen runtime dependencies and executed on a single GPU; the dependency set is specified by the environment file in the root directory of the anonymous repository. The main software and hardware environment is summarized in \cref{tab:appendix_environment}. The graph-centric results in the main paper are produced through a unified training entry that handles data loading, training, validation, and testing.

\begin{table}[t]
\centering
\small
\setlength{\tabcolsep}{5pt}
\renewcommand{\arraystretch}{1.12}
\begin{tabular}{ll}
\toprule
\textbf{Item} & \textbf{Specification} \\
\midrule
Operating system & Ubuntu 22.04.5 LTS \\
Python & 3.10.18 \\
PyTorch & 2.8.0 \\
CUDA runtime & 12.8 \\
Graph libraries & PyG 2.6.1, DGL 2.4.0 \\
GPU & NVIDIA RTX PRO 6000 Blackwell Server Edition \\
GPU memory & 97{,}887 MiB \\
\bottomrule
\end{tabular}
\caption{\textbf{Main software and hardware environment used in the experiments.}}
\label{tab:appendix_environment}
\end{table}

For node classification, the logs record the training loss, validation and test accuracy / macro-F1, as well as routing statistics on observed edges at the epoch level. For link prediction, the logs record the training loss, validation MRR, and Hits@$K$, and report the test performance at the checkpoint with the best validation result. The complete environment file and experimental configurations can be found in the anonymous repository: \href{https://anonymous.4open.science/r/RoleMAG-7EE0/}{anonymous.4open.science/r/RoleMAG-7EE0/}.

\section{Reproducibility Commands and Output Contract}
\label{app:repro}

\subsection{Task Configuration Interface}
\label{app:repro_scope}

The graph-centric experiments underlying the main-paper results include node classification on Toys, node classification on RedditS, and link prediction on Bili\_Dance. The corresponding data splits, input construction, evaluation metrics, and training setup are described in \cref{app:datasets_splits,app:preprocess,app:benchmark_protocol,app:rolemag_impl_setup,app:optimization,app:environment_commands}. A single run is denoted by
\[
\mathcal{C}=(\mathcal{T},\mathcal{D},\mathcal{M},\Theta),
\]
where $\mathcal{T}$ denotes the task type, $\mathcal{D}$ denotes the dataset instance, $\mathcal{M}$ denotes the model identifier, and $\Theta$ collects the remaining training and inference hyperparameters. RoleMAG follows the unified graph-centric entrypoint of OpenMAG~\cite{wan2026openmag}, with \texttt{src/main.py} as the invocation interface. At runtime, $\mathcal{C}$ is instantiated by Hydra overrides on the three core fields \texttt{task}, \texttt{dataset}, and \texttt{model}, after which data preparation, model construction, training, validation, and testing are executed in sequence.

\subsection{Canonical Invocation}
\label{app:repro_commands}

The experiment dependencies are specified by the environment file at the repository root, and the execution entrypoint is written as
\[
\texttt{python src/main.py}.
\]
Within the scope of the main paper, the three graph-centric experiments correspond to the following command templates:

\begin{quote}
\small\ttfamily
python src/main.py task=nc dataset=toys model=rolemag

python src/main.py task=nc dataset=reddits model=rolemag

python src/main.py task=lp dataset=bili\_dance model=rolemag
\end{quote}

These three commands correspond to the two node classification tasks and the single link prediction task, respectively. All remaining hyperparameters inherit the setup specified in \cref{app:setup}, including batch size, sampling depth, optimizer, learning rate, weight decay, negative sampling scale, and the RoleMAG-specific configurations. In a more compact form, the same interface can be written as
\[
\texttt{python src/main.py task=}\mathcal{T}\ \texttt{dataset=}\mathcal{D}\ \texttt{model=rolemag}\ \Theta.
\]

\subsection{Task-level Output Semantics}
\label{app:repro_outputs}

For node classification, the basic logging unit is the epoch-level record, which contains at least the training loss \texttt{Loss}, the validation and test \texttt{Acc}, and the corresponding \texttt{Macro-F1}. When routing diagnostics are additionally recorded, the log may also include role posteriors, gate weights, and uncertainty statistics, characterizing how the three neighborhood signal types, namely shared, complementary, and heterophily, are allocated during training. A node classification run therefore yields two classes of outputs: the trajectories of training loss and validation metrics, and the final classification performance at the reported test point.

For link prediction, the training stage outputs epoch-level \texttt{Train Loss}, while the validation and test stages report \texttt{MRR} and \texttt{Hits@1/3/10}. The test results are reported from the checkpoint selected by the best validation \texttt{MRR}. Accordingly, the output structure of link prediction consists of a training-loss trajectory together with the associated ranking metrics.

At the level of output entities, what is fixed here is the field semantics rather than any specific filename. Once the configuration $\mathcal{C}$ is determined, the training process corresponds to a unique task context and metric stream; whether the logs are printed to standard output, written to the run directory, or further aggregated into result files depends on the saving configuration.

\subsection{Implementation Scope}
\label{app:repro_nonclaims}

The RoleMAG implementation in the anonymous repository is integrated into the unified benchmark training entrypoint and executes the corresponding experiments under the task scope, configuration interface, and output fields specified in this paper. The command interface, task configuration, and output semantics described above are all grounded in this execution entrypoint. The remaining method definitions and complexity discussion are provided in \cref{app:method_spec,app:complexity_discussion}.

\section{More Experiments}
\label{app:optional}

This section supplements the main paper with two verified sets of additional results, namely extra node classification results and hyperparameter sensitivity analysis. Unless otherwise specified, all experiments follow the data splits and evaluation protocols described in \cref{app:setup,app:benchmark_protocol}. As in the main paper, all metrics are reported in percentage.

\subsection{Additional Node Classification Results}
\label{sec:appendix-e1}

We report additional node classification results on two datasets beyond the main-paper triad. For readability, we retain the same strong baseline subset as in the main paper, namely DGF, LGMRec, and Graph4MM. The complete benchmark protocol is described in \cref{app:benchmark_protocol}.

\begin{table*}[t]
\centering
\small
\caption{\textbf{Additional node classification results} on four MAG datasets. All metrics are reported in \%.}
\label{tab:appendix-additional-nc}
\resizebox{\textwidth}{!}{
\begin{tabular}{lccccccccc}
\toprule
Dataset & \multicolumn{2}{c}{RoleMAG} & \multicolumn{2}{c}{DGF} & \multicolumn{2}{c}{LGMRec} & \multicolumn{2}{c}{Graph4MM} \\
\cmidrule(lr){2-3}\cmidrule(lr){4-5}\cmidrule(lr){6-7}\cmidrule(lr){8-9}
& Acc & F1 & Acc & F1 & Acc & F1 & Acc & F1 \\
\midrule
Toys        & 78.83 & 75.49 & 77.68 & 71.46 & 78.81 & 72.31 & \textbf{78.91} & \textbf{75.62} \\
RedditS     & \textbf{92.57} & \textbf{87.04} & 91.73 & 82.63 & 92.36 & 86.10 & 92.34 & 86.31 \\
Grocery     & 83.45 & 73.56 & 83.02 & 72.56 & \textbf{84.12} & \textbf{76.83} & 83.57 & 75.92 \\
Bili\_Dance & 44.82 & 39.14 & 41.73 & 36.21 & 43.90 & 37.88 & \textbf{45.21} & \textbf{39.87} \\
\bottomrule
\end{tabular}}
\end{table*}

The additional node classification results follow a pattern similar to that in the main paper. RoleMAG remains competitive on the extra datasets, while the strongest baseline varies across benchmarks. In particular, LGMRec attains the best result on Grocery, whereas Graph4MM performs best on Bili\_Dance. These results indicate that the effect of role-aware propagation depends on the dataset characteristics and the local neighborhood structure.

\subsection{Hyperparameter Sensitivity}
\label{sec:appendix-e3}

We further examine the influence of the TopK neighborhood size, the number of query tokens, the temperature parameter $\tau$, and the auxiliary loss weights on RedditS node classification. The corresponding results are summarized in \cref{tab:appendix-hparam-main,tab:appendix-hparam-loss}. The default setting used in the main experiments is $K=16$, $Q=4$, $\tau=0.07$, $\lambda_{\mathrm{qca}}=0.5$, and $\lambda_{\mathrm{bal}}=0.1$.

\begin{table}[t]
\centering
\small
\caption{\textbf{Hyperparameter sensitivity of structural interaction parameters} on RedditS node classification. All metrics are reported in \%.}
\label{tab:appendix-hparam-main}
\begin{tabular}{lccc}
\toprule
Sweep & Value & Acc & F1 \\
\midrule
TopK & 8  & 93.49 & 87.10 \\
TopK & 16 & 92.92 & 86.25 \\
TopK & 32 & 93.14 & 86.68 \\
\midrule
Queries & 2 & 93.81 & 87.27 \\
Queries & 4 & 92.64 & 85.75 \\
Queries & 8 & 93.14 & 86.60 \\
\midrule
$\tau$ & 0.03 & 92.92 & 86.67 \\
$\tau$ & 0.07 & 92.39 & 86.35 \\
$\tau$ & 0.15 & 92.96 & 86.36 \\
\bottomrule
\end{tabular}
\end{table}

\begin{table}[t]
\centering
\small
\caption{\textbf{Hyperparameter sensitivity of auxiliary objective weights} on RedditS node classification. All metrics are reported in \%.}
\label{tab:appendix-hparam-loss}
\begin{tabular}{lccc}
\toprule
Sweep & Value & Acc & F1 \\
\midrule
$\lambda_{\mathrm{qca}}$ & 0.0 & 92.74 & 85.79 \\
$\lambda_{\mathrm{qca}}$ & 0.2 & 92.70 & 86.20 \\
$\lambda_{\mathrm{qca}}$ & 0.5 & 93.21 & 86.07 \\
$\lambda_{\mathrm{qca}}$ & 1.0 & 92.67 & 86.70 \\
\midrule
$\lambda_{\mathrm{bal}}$ & 0.0  & 92.61 & 85.56 \\
$\lambda_{\mathrm{bal}}$ & 0.05 & 93.11 & 86.69 \\
$\lambda_{\mathrm{bal}}$ & 0.1  & 91.98 & 85.24 \\
$\lambda_{\mathrm{bal}}$ & 0.2  & 93.33 & 86.65 \\
\bottomrule
\end{tabular}
\end{table}

Overall, the sensitivity results remain within a relatively narrow range over the tested settings. The influence of $K$ and the number of queries is moderate, and the tested values of $\tau$ yield comparable performance. The auxiliary weights $\lambda_{\mathrm{qca}}$ and $\lambda_{\mathrm{bal}}$ also affect the final metrics, which supports treating them as tunable regularization coefficients rather than fixed constants.

\section{Limitations and Disclosure}
\label{app:limits}

\zh{
\subsection{Scope and Assumptions}
\label{app:limits_scope}

RoleMAG, as instantiated in the main paper and the preceding appendix sections, follows the graph-centric multimodal learning setting and is evaluated under the OpenMAG benchmark protocol on two node classification tasks and one link prediction task~\cite{wan2026openmag}. Its task interface can be written as
\[
(\mathcal{G},\mathbf{X}^{T},\mathbf{X}^{I})
\longrightarrow
\psi_{\mathrm{task}}(\mathbf{Z}),
\]
where $\mathcal{G}$ denotes the observed graph, $\mathbf{X}^{T}$ and $\mathbf{X}^{I}$ denote the frozen text and image node representations, and $\mathbf{Z}$ is produced by the role-aware propagation and fusion process defined in the main paper; see \cref{app:method_spec,app:pseudocode,app:setup,app:repro}. The experimental scope discussed in this section rests on three basic assumptions. First, the graph structure is treated as an observed relation during both training and evaluation. Second, the bimodal node representations are aligned to a unified dimension through an offline encoding pipeline. Third, downstream learning focuses on neighborhood role decomposition and propagation organization rather than end-to-end updates of the underlying multimodal encoders.

Under this setting, RoleMAG models three types of local neighborhood behavior. The shared channel captures cross-modal consistent support, the complementary channel models directional cross-modal completion, and the heterophily channel preserves heterophilous relations in an independent high-frequency interaction path. Accordingly, the present study covers the problem family defined by fixed bimodal features, a given local graph structure, and graph-centric downstream tasks. Dynamic graphs, temporally evolving graphs, joint propagation over more than two modalities, and end-to-end co-optimization with generative multimodal models all fall outside the current experimental scope.

\subsection{Limitations and Failure Boundaries}
\label{app:limits_failure}

RoleMAG builds its role decomposition on bimodal availability factors and edge-level structural statistics. As a result, the current definitions of the router and the complementary directions are tied to the two-modality interface $\mathcal{M}=\{T,I\}$. If the number of modalities increases, the shared/complementary/heterophilous decomposition may still remain meaningful, but the parameterization of complementary directions, query routing, and role balancing would need to be redesigned. A direct reuse of the present bimodal factorization is unlikely to cover more general multimodal coupling patterns.

The current propagation backbone operates only on the observed edges defined by the task and relies on frozen node representations together with local neighborhood sampling to build mini-batch subgraphs; see \cref{app:preprocess,app:optimization}. Its effectiveness is therefore constrained by two input conditions. On the one hand, the graph structure must provide sufficiently stable local support for the target task. On the other hand, the offline multimodal representations must retain enough semantic separability. When the observed graph suffers from severe missing edges, relation-semantic drift, abrupt temporal structural changes, or simultaneous degradation of both modalities, the estimated role posteriors and the subsequent division of expert responsibilities can both deteriorate. RoleMAG changes how neighborhoods participate in propagation; it does not replace the quality of the underlying features themselves.

From a computational perspective, RoleMAG does not compress all neighbors into a single propagation path. Instead, it explicitly introduces directional complementary completion, heterophily-aware filtering, and node-wise gating. This design improves the expressiveness of propagation, but it also creates additional computation and hyperparameter coupling. The TopK size, the number of queries, and the attention-bias strength in the complementary channel, together with the polynomial coefficients in the heterophily channel, jointly affect training stability and efficiency. The efficiency analysis in the main paper already characterizes the resulting overhead. The design goal here is not to minimize computation, but to model finer-grained role-aware propagation under a unified benchmark protocol.

The experiments covered in the present appendix focus on graph-centric node classification and link prediction. Accordingly, task families such as retrieval, generation, instruction tuning, and larger-scale foundation-model adaptation are not developed in this supplementary material. These directions are related to the broader idea of role-aware propagation, but they require different data construction pipelines, training interfaces, and evaluation protocols, and are therefore outside the direct scope of the conclusions reported here.

\subsection{Disclosure and Anonymity}
\label{app:limits_disclosure}

The main-paper results, the command interface in the appendix, and the accompanying experimental descriptions are all grounded in the same anonymous benchmark pipeline. The task entrypoint and configuration interface are given in \cref{app:repro}, the experimental protocol follows the graph-centric setting of OpenMAG, and RoleMAG is integrated into the shared training and evaluation workflow through a unified entrypoint~\cite{wan2026openmag}. Accordingly, the method modules, task scope, data splits, optimization setup, and output fields described in Appendix B--F and Appendix H all refer to the same set of experiments.

The disclosed resource boundary is as follows. Training and evaluation are both performed on frozen bimodal features. The main software and hardware environment, dependency specification, and execution commands are described in \cref{app:environment_commands,app:repro,app:checklist}. The data sources, task splits, and metric definitions inherit the public benchmark protocol. The additional computation introduced by RoleMAG mainly comes from four components: the role router, complementary retrieval, heterophily filtering, and gating fusion. The supplementary material therefore fixes the task protocol, model interface, and output semantics around this scope.

The anonymity constraint applies to both the supplementary document and the code assets. The repository link, command templates, environment descriptions, figure and table captions, and cross-references in the appendix all remain anonymous. They do not contain author names, institutional names, personal webpages, non-anonymous project addresses, identity-bearing file metadata, or traceable local paths. The same constraint applies to the paper PDF, the appendix, the anonymous repository, and the exposed configuration assets associated with it.
}

\section{Reproducibility Checklist}
\label{app:checklist}

\zh{
The following table summarizes the execution interface, configuration conventions, and output fields underlying the graph-centric main-paper results.

\begin{table*}[t]
\centering
\small
\setlength{\tabcolsep}{5pt}
\renewcommand{\arraystretch}{1.12}
\begin{tabular}{p{0.25\textwidth} p{0.43\textwidth} p{0.24\textwidth}}
\toprule
\textbf{Reproducibility Item} & \textbf{Specification} & \textbf{Reference} \\
\midrule
Execution entrypoint & The graph-centric experiments are launched through \texttt{python src/main.py}, where the three fields \texttt{task}, \texttt{dataset}, and \texttt{model} jointly determine a run configuration. & \cref{app:environment_commands,app:repro_commands} \\
Environment specification & The dependency set is frozen by the environment file at the repository root, and the main software and hardware environment is listed in the environment table. & \cref{app:environment_commands,tab:appendix_environment} \\
Task scope & The main paper covers two node classification tasks (Toys and RedditS) and one link prediction task (Bili\_Dance). & \cref{app:datasets_splits,app:repro_scope} \\
Data split protocol & Node classification uses a $60/20/20$ node split, while link prediction uses a $70/10/20$ edge split and is evaluated with ranking metrics. & \cref{app:datasets_splits,app:benchmark_protocol} \\
Input construction & Node inputs are constructed from frozen multimodal embeddings together with task-specific graph preprocessing. & \cref{app:preprocess} \\
Model instantiation & RoleMAG is integrated into the training and evaluation workflow through the unified benchmark interface and instantiated along the router-expert-fusion pipeline specified in this paper. & \cref{app:rolemag_impl_setup,app:repro_nonclaims} \\
Core modules & The execution path of RoleMAG includes the edge-role router, the shared/complementary/heterophily experts, residual gating fusion, and the associated optimization objectives. & \cref{app:edge_role_router,app:shared_expert,app:complementary_expert,app:heterophily_expert,app:fusion_objectives} \\
Optimization setup & The optimizer, learning rate, weight decay, dropout, training epochs, and task-specific batch configurations all follow the experimental setup given in Appendix~D. & \cref{app:optimization,tab:appendix_rolemag_hparams} \\
Node classification outputs & Each run reports epoch-level loss together with validation and test classification metrics, and may additionally include routing statistics for diagnosis. & \cref{app:repro_outputs} \\
Link prediction outputs & Each run reports the training loss and the ranking metrics on the validation and test sets, including \texttt{MRR} and \texttt{Hits@K}. & \cref{app:repro_outputs} \\
Anonymous repository contents & The anonymous repository provides the experimental interface, dependency environment, and training entrypoint described in this paper. & \cref{app:environment_commands,app:repro} \\
Main experimental protocol & The graph-centric setting documented in this supplementary material covers the same task scope, data splits, and evaluation protocol as the main-paper results. & \cref{app:datasets_splits,app:benchmark_protocol,app:repro_nonclaims} \\
Cross-section consistency & The command interface, the detailed experimental setup, and the method specification all correspond to the same task configuration, model interface, and output semantics. & \cref{app:method_spec,app:setup,app:repro} \\
\bottomrule
\end{tabular}
\caption{\textbf{RoleMAG supplementary reproducibility checklist.} The listed items specify the execution interface, configuration conventions, and output fields underlying the graph-centric experiments in the main paper.}
\label{tab:repro_checklist}
\end{table*}

The graph-centric main-paper results are jointly determined by the execution entrypoint, environment dependencies, task scope, data splits, input construction, the RoleMAG instantiation, the optimization setup, and the task-level output fields. These entities align one by one with the method specification and the experimental setup given in the preceding sections.

\clearpage
}


\end{document}